\documentclass[10pt,journal,compsoc]{IEEEtran}
\usepackage[colorlinks,linkcolor=black, citecolor=black]{hyperref}
\usepackage{color}
\usepackage{graphicx}
\usepackage{amssymb}
\usepackage{amsfonts}
\usepackage{mathrsfs}
\usepackage{url}
\usepackage{amsmath}
\usepackage{algorithm}
\usepackage{algorithmic}
\usepackage{multirow}
\usepackage{booktabs}
\usepackage[table]{xcolor}
\usepackage{setspace}
\usepackage{siunitx}
\usepackage{upgreek}
\usepackage{makecell}
\usepackage{multirow}
\usepackage{soul}

\usepackage{bbding}
\definecolor{mygray}{gray}{.9}
\definecolor{babyblueeyes}{rgb}{0.7, 0.8, 1}

\renewcommand{\arraystretch}{1}

\ifCLASSOPTIONcompsoc
  \usepackage[nocompress]{cite}
\else
  \usepackage{cite}
\fi
\ifCLASSINFOpdf
\else
\fi
\begin{document}
%
\title{HSIGene: A Foundation Model for Hyperspectral Image Generation}

\author{Li Pang, Xiangyong Cao, Datao Tang, Shuang Xu, Xueru Bai, Feng Zhou, Deyu Meng
\thanks{Li Pang, Datao Tang and Xiangyong Cao are with the School of Computer Science and Technology and the Ministry of Education Key Lab for Intelligent Networks and Network Security, Xi’an Jiaotong University, Xi’an, Shaanxi 710049, China (Email: caoxiangyong@xjtu.edu.cn) (\textit{Corresponding author: Xiangyong Cao}).}
\thanks{Shuang Xu is with the School of Mathematics and Statistics, Northwestern Polytechnical University, Xi’an, Shaanxi 710021, China. }
\thanks{Xueru Bai is with the National Laboratory of Radar Signal Processing, Xidian University, Xi’an 710071, China.}
\thanks{Feng Zhou is with the Key Laboratory of Electronic Information Countermeasure and Simulation of the Education Ministry of China, Xidian University, Xi’an 710071, China.}
\thanks{Deyu Meng is with the School of Mathematics and Statistics and the Ministry of Education Key Laboratory of Intelligent Networks and Network Security, Xi’an Jiaotong University, Xi’an, Shaanxi 710049, China, and also with Pazhou Laboratory (Huangpu), Guangzhou, Guangdong 510555, China.}}

%
%

\markboth{Journal of \LaTeX\ Class Files,~Vol.~14, No.~8, August~2015}%
{Shell \MakeLowercase{\textit{et al.}}: Bare Demo of IEEEtran.cls for Computer Society Journals}
%



\IEEEtitleabstractindextext{%
\begin{abstract}
Hyperspectral image (HSI) plays a vital role in various fields such as agriculture and environmental monitoring. However, due to the expensive acquisition cost, the number of hyperspectral images is limited, degenerating the performance of downstream tasks. Although some recent studies have attempted to employ diffusion models to synthesize HSIs, they still struggle with the scarcity of HSIs, affecting the reliability and diversity of the generated images. Some studies propose to incorporate multi-modal data to enhance spatial diversity, but spectral fidelity cannot be ensured. In addition, existing HSI synthesis models are typically uncontrollable or only support single-condition control, limiting their ability to generate accurate and reliable HSIs. To alleviate these issues, we propose HSIGene, a novel HSI generation foundation model which is based on latent diffusion and supports multi-condition control, allowing for more precise and reliable HSI generation. To enhance the spatial diversity of the training data while preserving spectral fidelity, we propose a new data augmentation method based on spatial super-resolution, in which HSIs are upscaled first, and thus abundant training patches could be obtained by cropping the high-resolution HSIs. In addition, to improve the perceptual quality of the augmented data, we introduce a novel two-stage HSI super-resolution framework, which first applies RGB bands super-resolution and then utilizes our proposed Rectangular Guided Attention Network (RGAN) for guided HSI super-resolution. Experiments demonstrate that the proposed model is capable of generating a vast quantity of realistic HSIs for downstream tasks such as denoising and super-resolution. The code and models are available at \href{https://github.com/LiPang/HSIGene}{\textcolor{blue}{https://github.com/LiPang/HSIGene}}.
\end{abstract}

}

\maketitle

\IEEEdisplaynontitleabstractindextext

%
\IEEEpeerreviewmaketitle

\IEEEraisesectionheading{\section{Introduction}\label{sec:introduction}}
\IEEEPARstart{H}{yperspectral} image (HSI) is captured across a continuous range of wavelengths, providing detailed information on the spectral characteristics of different materials. HSI plays a crucial role in various applications such as remote sensing~\cite{thenkabail2016hyperspectral, manolakis2016hyperspectral}, medical~\cite{lu2014medical, calin2014hyperspectral} and agriculture~\cite{lu2020recent, dale2013hyperspectral}. In recent years, with the rapid advancement of artificial intelligence, deep learning (DL) techniques have been widely adopted across various HSI applications, including classification~\cite{audebert2019deep,cao2020hyperspectral}, denoising~\cite{dong2019deep, shi2021hyperspectral, pang2024hir}, super-resolution~\cite{arun2020cnn, hu2021hyperspectral} and so on. However, owing to the high cost associated with hyperspectral data acquisition, the number of high-quality HSIs is limited, posing significant challenges to the application of DL in HSI processing tasks.

To alleviate the limitation of the scarcity of HSIs, a promising alternative approach involves the synthetic generation of HSIs using deep generative models. Deep generative models, such as Variational Autoencoders (VAEs)~\cite{kingma2013auto}, Generative Adversarial Networks (GANs)~\cite{goodfellow2014generative}, and Diffusion Models (DMs)~\cite{ho2020denoising, song2020denoising}, have shown remarkable success in generating realistic data across various domains. Among them, diffusion models have gained increasing popularity, since they have achieved excellent generation results and can avoid some of the common drawbacks of VAEs and GANs, such as mode collapse in GANs and posterior collapse in VAEs. Furthermore, the technique of Latent Diffusion Models (LDMs)~\cite{rombach2022high} combines the power of diffusion models with the efficiency of latent space representations, significantly reducing computational burden while maintaining desirable visual fidelity. Owing to its impressive synthesis capability, applying generative models to hyperspectral imaging is an area of growing interest.

\begin{figure}[h]
  \centering
  \includegraphics[width=1\linewidth]{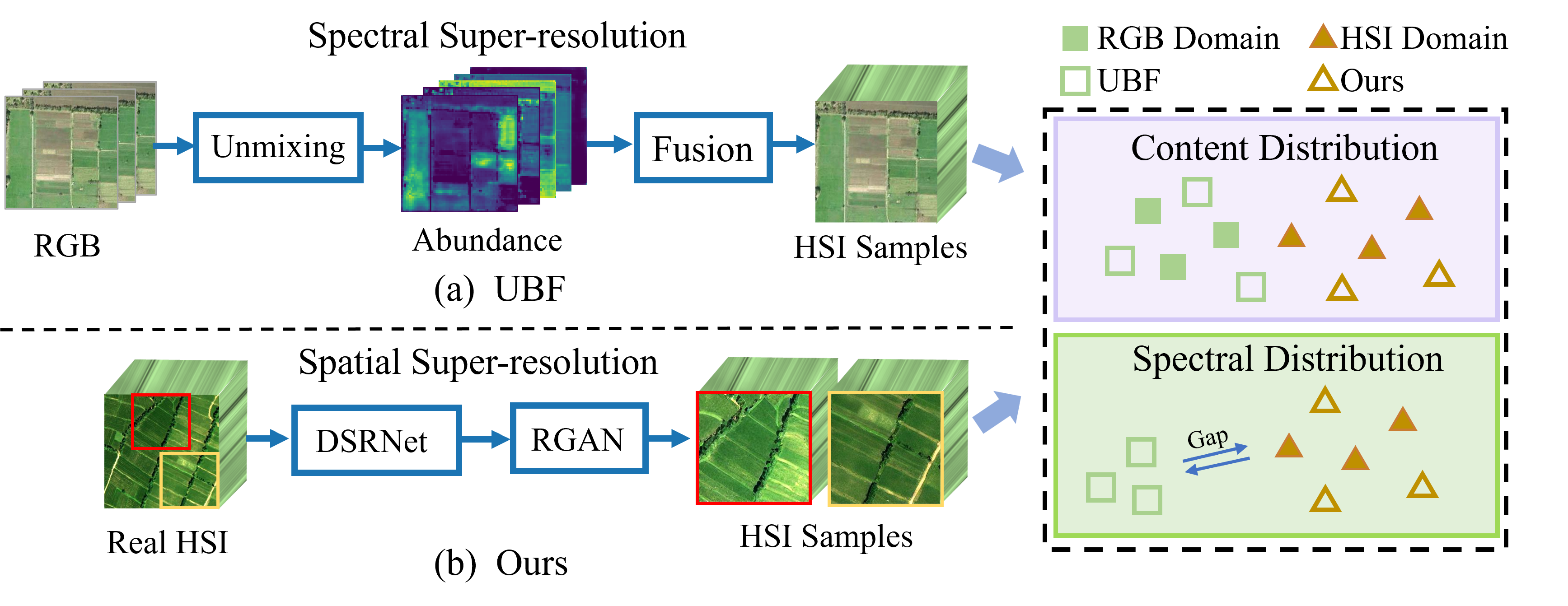}
  \caption{A schematic comparison of the existing HSI augmentation methods. UBF~\cite{yu2024unmixing} alleviates the issue of the data sacrifice by performing spectral super-resolution on external RGB images, resulting in data with similar content to RGB images but with less authentic spectral profiles. In contrast, our method performs spatial super-resolution on existing real HSIs, ensuring that both the content and spectral distribution are consistent with real HSIs.}\label{fig:image1}
\end{figure}

\begin{figure*}[ht]
  \centering
  \includegraphics[width=0.7\textwidth]{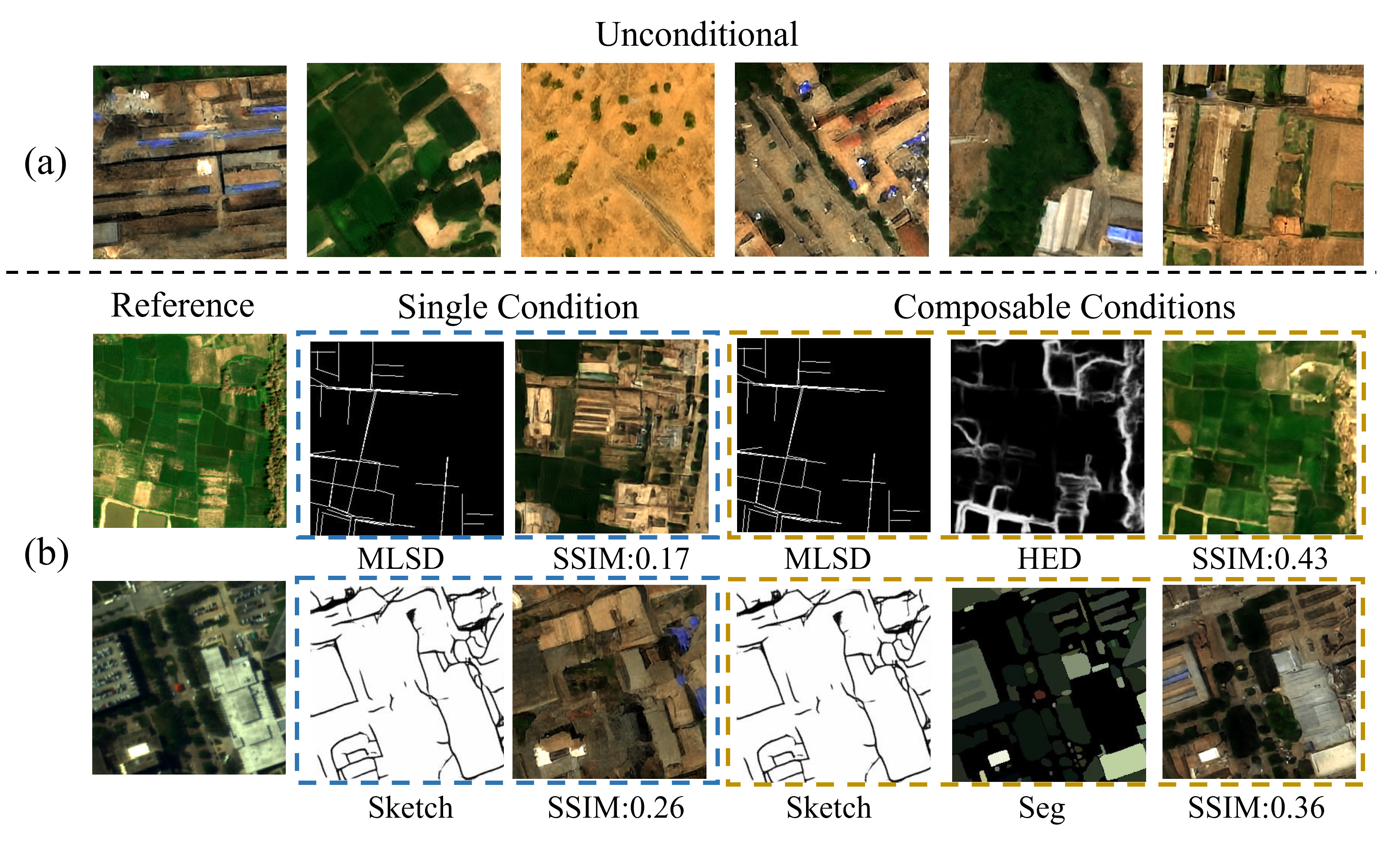}
  \caption{Visualization results of our proposed HSIGene in different situations. When provided with more conditions, they can complement each other to achieve more accurate generation. (a) Unconditional generation results. (b) Generation results under single condition and multiple conditions. }\label{fig:syn1}
\end{figure*}



Recently, some studies have attempted to synthesize HSIs with diffusion models~\cite{yu2024unmixing, yu2024unmixdiff}. One representative work, Unmixing before Fusion (UBF)~\cite{yu2024unmixing}, synthesizes hyperspectral data in the abundance domain and incorporates RGB images to increase the diversity of training HSIs. Specifically, an unmixing network is first trained on real HSIs to estimate endmembers and abundances. After that, as shown in Fig.~\ref{fig:image1}(a), the abundances of external RGB images are inferred utilizing the unmixing network. Finally, a fusion-based generative model is trained to synthesize abundance maps, and new HSI samples are generated by combining the synthesized abundance maps with the estimated endmembers. Although the model can generate HSIs, due to the inherent spectral and content differences between hyperspectral and RGB data, the fidelity of the spectral profiles of the generated HSIs cannot be guaranteed. Instead, UnmixDiff~\cite{yu2024unmixdiff}, directly trains the generative model with the abundance maps of real HSIs. However, UnmixDiff still suffers from the scarcity of high-quality HSIs and limited model size, limiting the diversity of the generated HSIs. In addition, UBF and Unmixdiff both synthesize abundance maps instead of HSIs directly, resulting in the quality of the generated spectra being highly limited by the performance of the unmixing network. Moreover, only a single condition can be incorporated to control the synthesis process, and thus Unmixdiff cannot guarantee the accuracy and reliability of the generated HSIs.

\begin{table}[t]
\caption{Comparison between our model and existing HSI synthesis models.}
\label{tab:comparison}
\renewcommand\arraystretch{1.1}
\resizebox{\linewidth}{!}{
\begin{tabular}{lccccc}
\toprule
Method    & Dataset        & Latent DM & Controllable & Conditions & \makecell{Composable \\ control} \\ \midrule
UBF~\cite{yu2024unmixing}       & Synthetic      & \XSolidBrush         & \XSolidBrush            & 0          & \XSolidBrush                  \\
UnmixDiff~\cite{yu2024unmixdiff} & Real           & \XSolidBrush         & \Checkmark            & 2          & \XSolidBrush                  \\
Ours      & Synthetic+Real & \Checkmark         & \Checkmark            & 6          & \Checkmark                  \\ \bottomrule
\end{tabular}
}
\end{table}

To alleviate these issues, we propose a novel HSI synthesis foundation model, HSIGene, which is based on latent diffusion models and supports multi-condition controllable generation. As illustrated in Fig.~\ref{fig:syn1}, our model is capable of generating HSIs both unconditionally and under the guidance of one or more conditions. Multiple conditions can be combined to provide more comprehensive information, resulting in more precise generation. To alleviate the scarcity of high-quality HSIs and enhance the spatial diversity of training samples, we perform spatial super-resolution (SR) on existing hyperspectral data as shown in Fig.~\ref{fig:image1}(b). Abundant HSI samples could be obtained by cropping the upscaled HSIs instead of the original real data. Since there are rich land covers inside a real HSI, the cropping patches of the upscaled HSIs exhibit various content, leading to better generation performance and generalization capabilities. Considering the challenges associated with real-world hyperspectral super-resolution (i.e. the lack of ground-truth high-resolution HSIs for training networks), we adopt a two-stage super-resolution framework that leverages the power of abundant high-resolution RGB images and diffusion models to improve the perceptual quality of the augmented data. Specifically, we first collect a large dataset of high-resolution RGB images that closely resemble real hyperspectral data by utilizing the latitude and longitude information. Utilizing the collected high-resolution RGB images, a diffusion model-based RGB super-resolution network (DSRNet) is trained and is then used to super-resolve the RGB bands of real HSIs, resulting in high-resolution RGB bands. Finally, with high-resolution RGB bands as guidance, a novel rectangular guided attention network (RGAN) is proposed to obtain high-resolution HSIs. A more detailed comparison of our model and existing works is demonstrated in Tab.~\ref{tab:comparison}.

Overall, our contributions can be summarized as follows.

\begin{itemize}
\item We propose a foundation model namely HSIGene for hyperspectral image generation. Our model can generate high-quality HSIs under various control conditions such as sketch and segmentation. As far as we know, HSIGene is the first and largest HSI generative model, supporting multiple control conditions.

\item We propose a new paradigm to alleviate the issue of limited HSI data availability by performing spatial super-resolution. With more diverse training samples, the generative capabilities and generalization performance of our model can be enhanced.

\item To improve the perceptual quality of augmented data, we propose a two-stage framework for HSI super-resolution including RGB bands super-resolution and RGB-guided HSI super-resolution. Additionally, a novel rectangular guided attention network (RGAN) is proposed to fully transfer the detailed supplementary information from the RGB modality to the HSI modality.

\item Experiment results on two downstream HSI tasks demonstrate that the synthetic HSI data could improve the performance and generalization ability of deep learning methods significantly, verifying the reliability of the proposed HSIGene to generate high-quality data for downstream tasks.

\end{itemize}

The rest of the paper is organized as follows: In Section~\ref{sec:related works}, we review the related work on the generative models, HSI synthesis and HSI super-resolution. Section~\ref{sec:method} presents our proposed HSIGene in detail, outlining the synthesis framework and the architecture of a guided super-resolution network. Section~\ref{sec:experiments} provides comprehensive experimental results and Section~\ref{sec:conclusion} summarizes our work.

\section{Related Works}
\label{sec:related works}
\subsection{Diffusion Model}
Recently, diffusion models~\cite{ho2020denoising, song2020denoising} have emerged as a powerful tool for generating samples from complex data distributions, with significant applications across various domains. Diffusion models gradually transform simple noise distributions into complex data distributions through a series of iterative denoising steps. Among these methods, Latent Diffusion Models (LDMs)~\cite{rombach2022high} have gained attention for their ability to operate in a compressed latent space, significantly reducing computational costs while maintaining high generation quality. Besides, the cross-attention layers in LDMs enable the models to generate content by conditioning on inputs such as text or bounding boxes. Over the past two years, various Controllable Diffusion Models (CDMs) have been proposed, enabling the generation of highly realistic images based on various inputs, such as text, sketches, or specific attributes. For example, T2I-Adapter~\cite{mou2024t2i} proposes a lightweight and efficient model to enhance the controllability of pre-trained text-to-image diffusion models without altering the original network structure. Uni-ControlNet~\cite{zhao2024uni} integrates diverse local and global controls into text-to-image diffusion models through the use of two additional adapters, offering flexible and composable control over image generation. There are also some generative works in the remote sensing field. DiffusionSat~\cite{khanna2023diffusionsat} employs a novel 3D ControlNet to enable a more flexible and high-quality generation of satellite images for various applications. CRS-Diff~\cite{tang2024crs} integrates the capabilities of diffusion models with advanced control mechanisms, supporting multiple control inputs, including text, metadata, and image conditions, to guide the generation process. However, HSI synthesis with multiple conditions remains unexplored.

\subsection{Hyperspectral Image Synthesis}
Owing to limited high-quality HSI data and the characteristics of high dimensionality, only a few methods that focus on HSI synthesis with deep generative models~\cite{yu2024unmixing, yu2024unmixdiff} are proposed. Unmixing before Fusion (UBF)~\cite{yu2024unmixing} introduces a novel paradigm for synthesizing HSIs by leveraging unmixing across multi-source data followed by fusion-based synthesis. In the method, an unmixing network is first trained on HSI data to extract endmembers and abundance maps. Using the trained unmixing model, the abundance from unpaired RGB data is inferred to train diffusion models. Finally, by fusing the estimated endmembers with the synthetic abundances generated by diffusion models, the approach effectively addresses the data scarcity issue in HSI research. Due to the inherent spectral and content differences between hyperspectral and RGB data, the model, while capable of generating HSIs, cannot ensure the fidelity of the spectral profiles of the generated images. UnmixDiff~\cite{yu2024unmixdiff}, similar to UBF, also generates HSIs by synthesizing abundance maps, which are then fused with estimated spectral endmembers to synthesize HSIs. While trained directly on real HSIs, UnmixDiff still suffers from the scarcity of high-quality HSIs and only supports generation with a single condition.

\subsection{Hyperspectral Image Super-resolution}
The majority of existing HSI-SR approaches are based on Convolutional Neural Networks (CNN) and Transformers. For example, GDRRN~\cite{li2018single} employs a grouped recursive module within a deep neural network to enhance the spatial resolution of HSIs while minimizing spectral distortion. MCNet~\cite{li2020mixed} utilizes a novel mixed convolutional module (MCM) that combines 2D and 3D convolutions to effectively extract spatial and spectral features for HSI super-resolution. Bi-3DQRNN~\cite{fu2021bidirectional} integrates a 3D convolutional module with a bidirectional quasi-recurrent pooling module to capture spatial-spectral structures and global spectral correlations. SSPSR~\cite{jiang2020learning} employs a group convolution strategy with shared parameters and a progressive upsampling framework for super-resolution of hyperspectral imagery. ESSAformer~\cite{zhang2023essaformer} incorporates a self-attention mechanism to further capture spatial-spectral information and long-range dependencies. However, these methods ignore the ill-posed nature of super-resolution tasks and tend to generate blurry outputs. Recently, a few approaches proposed to employ diffusion models to enhance the super-resolution performance. DMGASR~\cite{wang2024enhancing} integrates a Group-Autoencoder and a diffusion model to enhance the spatial resolution of HSIs while preserving spectral correlations. S2CycleDiff~\cite{qu2024s2cyclediff} leverages a conditional cycle-diffusion process and spatial/spectral guided pyramid denoising to enhance both spatial details and spectral accuracy. While these models demonstrate promising performance when applied to datasets similar to training data, such approaches suffer the scarcity of HSI data and could exhibit poor generalization in real-world settings.

\begin{table*}[ht]
\centering
\caption{Datasets used for training the HSI synthesis model.}
\label{tab:data}
\renewcommand\arraystretch{1.1}
\resizebox{0.85\textwidth}{!}{
\begin{tabular}{llllll}
\toprule
Name     & Spectral Range & Size                                & Bands & Device                                       & GSD  \\ \midrule
Xiongan  & 400-1000nm     & 3750$\times$1580                   & 250   & Airborne Multi-Modality Imaging Spectrometer & 0.5m \\
Chikusei & 343-1018nm     & 2517$\times$2335                   & 128   & Headwall Hyperspec-VNIR-C                    & 2.5m \\
DFC2013  & 380-1050nm     & 349$\times$1905                    & 144   & ITRES CASI-1500                              & 2.5m \\
DFC2018  & 380-1050nm     & 601$\times$2384                    & 48    & ITRES CASI 1500                              & 1m   \\
Heihe    & 380-1050nm     & Approx. 765$\times$512$\times$512  & 48    & CASI                                         & 1m   \\ \bottomrule
\end{tabular}
}
\end{table*}

\section{Methods}
\label{sec:method}
\subsection{Dataset Description}
\label{sec:data}

\subsubsection{HSI data collection}
In the process of compiling the training dataset, we encountered a variety of data sources with different qualities and characteristics. However, not all data are suitable for inclusion in our study. Specifically, we choose to exclude certain datasets due to their low resolution (e.g., Hyspecnet-11k~\cite{fuchs2023hyspecnet}, WHU-OHS~\cite{li2022whu}), high noise levels (e.g., Hyperion data\footnote{\url{https://www.usgs.gov/centers/eros/science/usgs-eros-archive-earth-observing-one-eo-1-hyperion}}), small size (e.g., Salinas\footnote{\url{https://www.ehu.eus/ccwintco/index.php/Hyperspectral_Remote_Sensing_Scenes}}), and the obscurity of semantic information. Finally, five HSI datasets, including Xiongan~\cite{yi2020aerial}, Chikusei~\cite{NYokoya2016}, DFC2013\footnote{\url{https://hyperspectral.ee.uh.edu/?page_id=459}}, DFC2018\footnote{\url{https://hyperspectral.ee.uh.edu/?page_id=1075}} and Heihe~\cite{HiWATER, li2017multiscale}, are employed as the training set of the generative models. A more detailed description of these datasets is provided in Table~\ref{tab:data}. However, the scarcity of high-quality HSIs might hinder the ability of our model to generate images from diverse conditions. Therefore, we propose a novel data augmentation strategy by performing spatial super-resolution to improve the generalization performance of our model, and more details are illustrated in Sec. \ref{sec:aug}.

\begin{figure*}[t]
  \centering
  \includegraphics[width=0.7\textwidth]{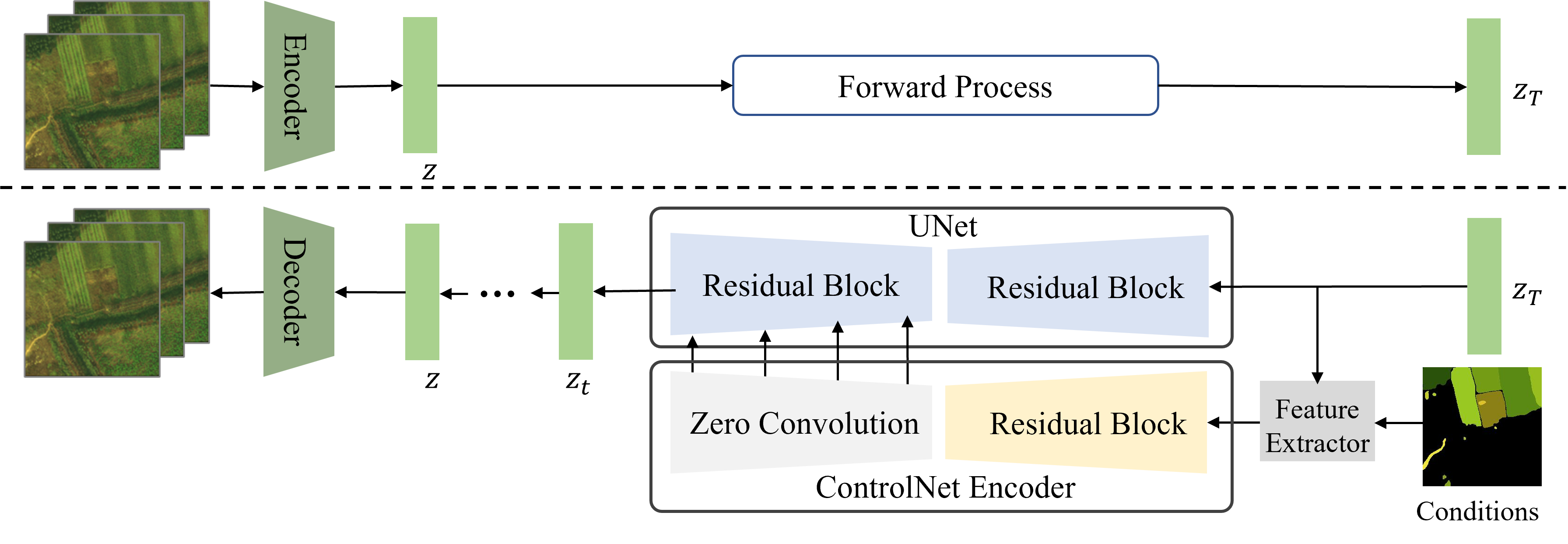}
  \caption{Overview of the generative model used in our work. The ControlNet encoder incorporates the information of the various conditions and hyperspectral images are generated with the diffusion process.}\label{fig:sd}
\end{figure*}

\subsubsection{Condition generation}
In our work, we consider six control conditions for HSI synthesis, including holistically nested edge detection (HED), segmentation, sketch, multiscale line segment detection (MLSD), content and text. We use the RGB bands of HSIs to generate conditions since most existing models are designed for RGB images. A more detailed description of the conditions is provided in the following.

\textbf{HED (Holistically-nested Edge Detection):} We employ the pre-trained model proposed in~\cite{xie2015holistically} to extract rich hierarchical representations that represent edge and object boundary.

\textbf{Segmentation:} The pre-trained segmentation model proposed in~\cite{kirillov2023segment} is utilized to generate the segmentation mask of HSI data.

\textbf{Sketch:} The sketch drawings of HSI data are obtained using the model proposed in~\cite{simo2016learning}, which offers a simplified representation that reduces an image to its essential lines and contours defining the object shapes within the image.

\textbf{MLSD (Multiscale Line Segment Detection):} We use the model proposed in~\cite{xu2021line} to achieve line segment detection of HSI data, providing straight paths between different endpoints.

\textbf{Content:} The image encoder in CLIP model~\cite{radford2021learning} is used to extract the global representation of the image content. The CLIP's ability to understand and respond to textual prompts makes it a versatile tool for image content extraction.

\textbf{Text:} To enable text-to-image generation, we annotate 1k training images with labels corresponding to the categories of the images. The labelling process involved assigning each image to one of four specific categories: farmland, city building, architecture, and wasteland. The labelled text enables the model to support category-guided synthesis.

\subsection{Controllable Generation Model}
\label{sec:crs}
For multi-condition generation, we adopt CRS-Diff proposed in~\cite{tang2024crs} as the generative model considering the highly accurate and controllable characteristics of the framework, and a brief introduction is provided in the following. As shown in Fig.~\ref{fig:sd}, the generative model consists of a VAE encoder, a UNet structure and a ControlNet. The pre-trained VAE encoder compresses the image into a latent space that is perceptually equivalent, and the diffusion process is operated within this latent space, which is computationally more efficient. More precisely, the encoder encodes image $x\in \mathbb{R}^{H \times W\times C}$ into latent representation $z\in \mathbb{R}^{h\times w\times c}$, and the decoder could reconstruct the image $x$ from the latent space. In the forward process, the latent representation $z$ is gradually added with noise over a series of timesteps, which eventually converges to a Gaussian distribution. The forward process can be described by the following:
\begin{equation}\label{eq:forward}
q(z_t \mid z_{t-1}) = \mathbb{N}(z_t; \sqrt{\alpha_t} z_{t-1}, (1 - \alpha_t) \mathbf{I}),
\end{equation}
where $z_t$ represents the noisy data at time step $t$ and $\alpha_t$ is a variance schedule parameter that controls the amount of noise added at each step. In the reverse process, with the low-resolution image as guidance, the model starts from the noise distribution and gradually removes noise to generate high-resolution samples. In our work, the sampling method proposed in Denoising Diffusion Implicit Models (DDIM)~\cite{song2020denoising} is adopted since the method allows for non-sequential steps in the sampling process, leading to significant speedups in generating samples. The reverse process is given by:
\begin{equation}\label{eq:reverse}
z_{t-1} = \sqrt{\bar{\alpha}_{t-1}} \left( \frac{z_t - \sqrt{1 - \bar{\alpha}_t} \hat{\epsilon}_t}{\sqrt{\bar{\alpha}_t}} \right) + \sqrt{1 - \bar{\alpha}_{t-1}} \cdot \hat{\epsilon}_t,
\end{equation}
where $\bar{\alpha}_t = \prod_{s=1}^t \alpha_s$ and $\hat{\epsilon}_t$ is the noise predicted by the model. In the training process, the model is trained to minimize the mean squared error (MSE) between the predicted noise and the actual noise added to the data. The loss function can be expressed as:
\begin{equation}\label{eq:diffusion}
L = \mathbb{E}_{t, z_0, \epsilon_t} \left[ \| \epsilon_t - \hat{\epsilon}_t \|_2^2 \right],
\end{equation}
where $\epsilon_t$ is the Gaussian noise added to the latent representation at time step $t$, $\hat{\epsilon}_t$ is the noise predicted by the model, and $z_0$ is the latent representation of the high-quality image. After training, the model is able to transform random noise into realistic high-resolution samples by sequential denoising at each step with multiple conditions such as sketch and segmentation.

The conditions are fed into the ControlNet encoder to ensure that the generated content aligns with the original low-resolution image. Specifically, the conditions are first fed into the feature extractor module, which consists of a series of convolutional layers to capture spatial hierarchies and semantic features. Then, the outputs from the zero-convolution modules are concatenated with those from the UNet in the diffusion model to effectively integrate the conditions, ensuring that the generated images reflect the multi-conditional information.

\subsection{HSI Data Augmentation}
\label{sec:aug}
\subsubsection{Overall framework}
\begin{figure}[t]
  \centering
  \includegraphics[width=0.9\linewidth]{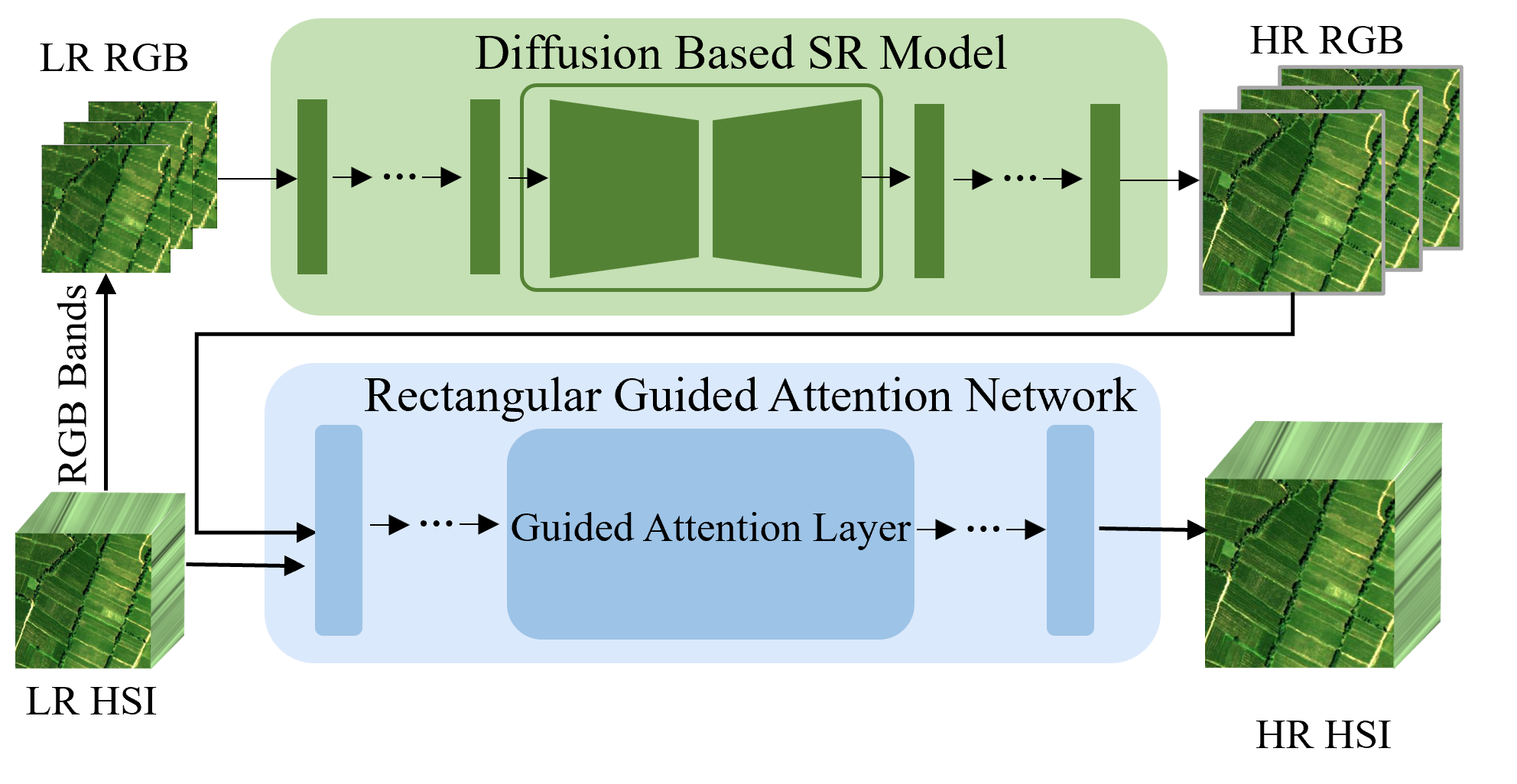}
  \caption{The overall framework of the proposed two-stage super-resolution framework. The RGB bands of HSIs are super-resolved first with a diffusion-based model (DSRNet). Then the high-resolution HSIs are obtained with a guided super-resolution network (RGAN) with the enhanced RGB bands as auxiliary prior information.}\label{fig:image2}
\end{figure}

To further improve the model's ability to handle various conditions, we propose to augment the training data with additional samples. The augmented data are expected to satisfy two constraints: Firstly, the augmented data should exhibit desirable perceptual quality. Secondly, the spectral distribution of the augmented data should closely match that of real HSIs. Therefore, we propose to perform spatial super-resolution for real HSIs with diffusion models as shown in Fig.~\ref{fig:image1}(b). By doing so, diverse training samples can be extracted by cropping the high-resolution HSIs. The augmented samples adhere to the two principles mentioned above as the super-resolution operation preserves spectral distribution and the diffusion models could ensure the high perceptual quality of the super-resolved results. To further improve the perceptual quality of the augmented data, we propose a novel two-stage framework for HSI super-resolution as shown in Fig.~\ref{fig:image2}, which takes advantage of abundant high-resolution remote sensing images to enhance super-resolution performance. In this framework, the RGB bands of HSIs are super-resolved first, and then the RGB details are utilized as a priori information to enhance the HSI super-resolution performance. Specifically, we first collect abundant high-resolution remote sensing images from Google Earth Engine~\cite{gorelick2017google} by matching their geographic coordinates with those of real HSIs, and thus the content of collected images is highly similar to real HSIs. Then, a diffusion-based RGB super-resolution network (DSRNet) is trained. This network enhances the resolution of the RGB bands of HSIs, producing high-resolution RGB bands. Finally, by using enhanced RGB bands as guidance, a novel Rectangular Guided Attention Network (RGAN) is proposed to generate high-resolution HSIs. More details of the networks are introduced in the following.

\begin{figure*}[ht]
  \centering
  \includegraphics[width=0.8\textwidth]{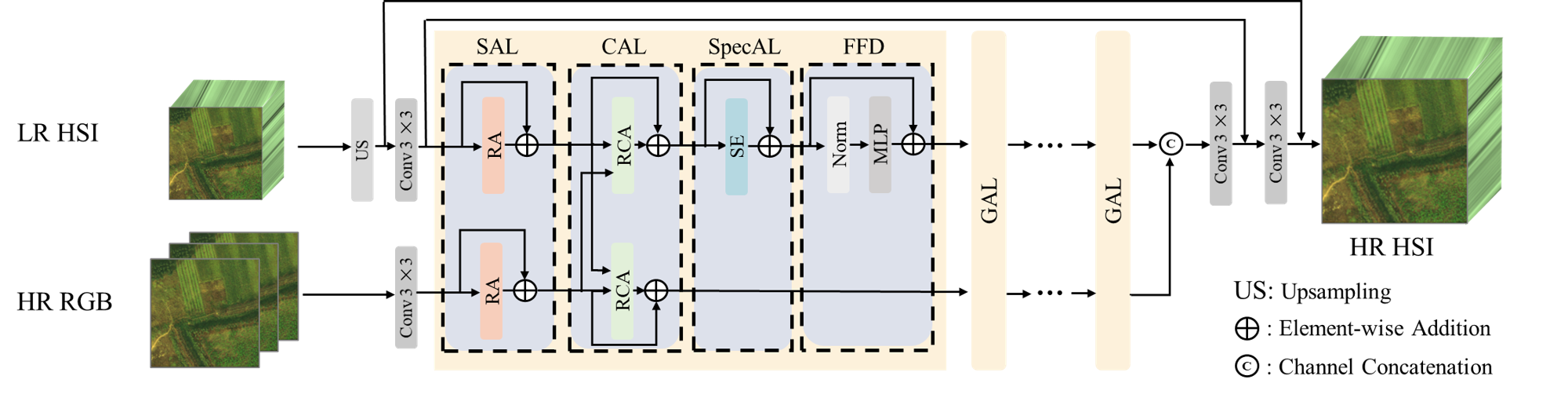}
  \caption{Overview of the proposed RGAN, which is composed of multiple guided attention layers (GALs). Each GAL is composed of self attention layer (SAL), cross attention layer (CAL), spectral attention layer (SpecAL) and feed forward layer (FFD). The network effectively transfers the fine details from the RGB modality into the hyperspectral modality, ensuring that the super-resolved hyperspectral images retain high fidelity and sharpness. }\label{fig:srn}
\end{figure*}

\subsubsection{Diffusion based super-resolution}
\label{sec:dbsr}
As diffusion models have achieved great success in the field of image generation, an increasing number of studies have applied diffusion models to image restoration tasks to enhance the perceptual quality of the restored results~\cite{li2022srdiff, gao2023implicit, wang2024sinsr}. In our work, we employ the diffusion-based super-resolution network (DSRNet) which is highly based on the framework introduced in Sec.~\ref{sec:crs} for RGB super-resolution. Specifically, we added an upsampling module before the feature extractor and regarded low-resolution images as conditions. The model is employed to obtain high-resolution RGB bands of HSIs, which provide image texture details for HSI super-resolution.

\begin{figure}[ht]
  \centering
  \includegraphics[width=1\linewidth]{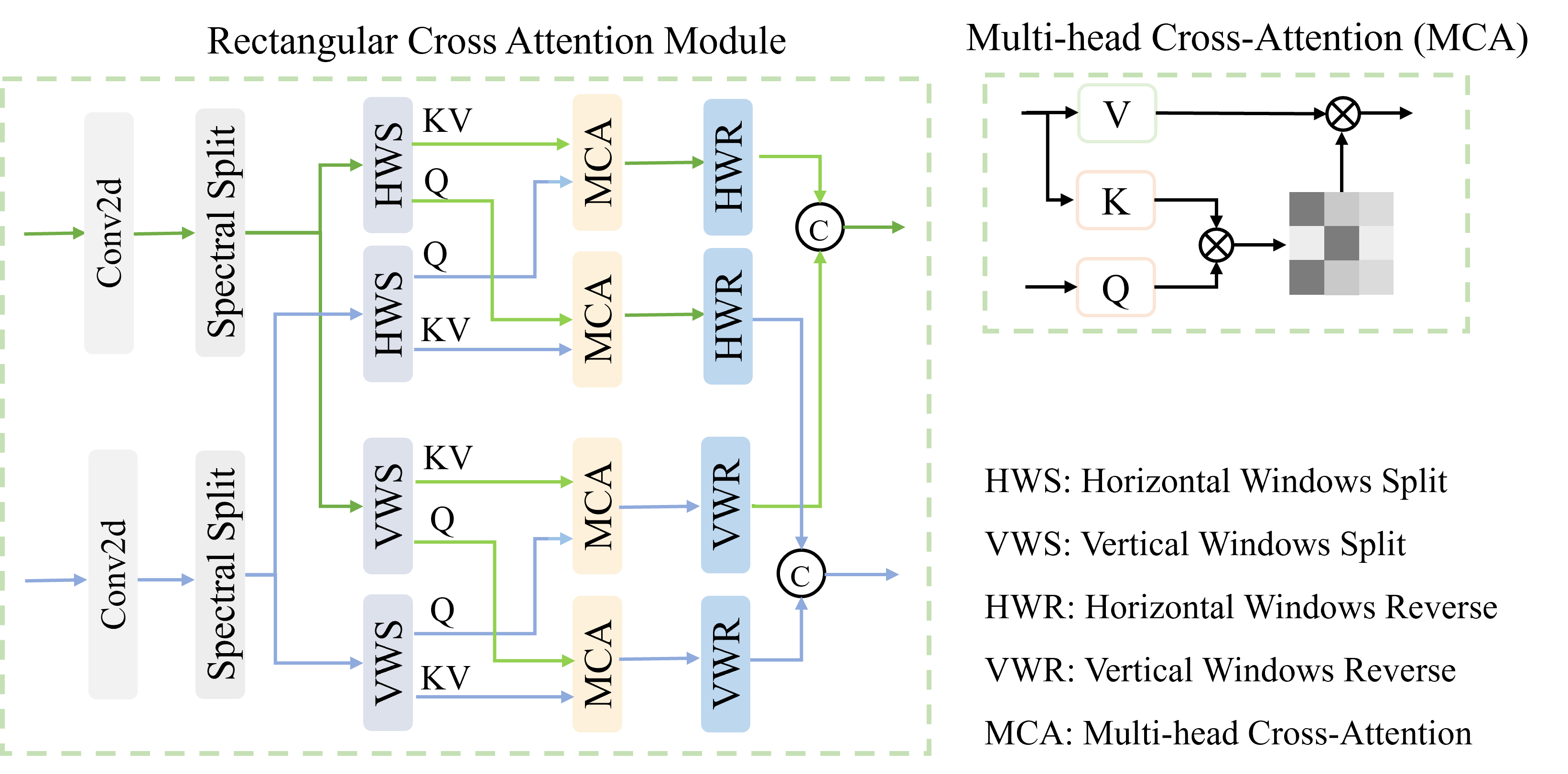}
  \caption{Illustration of the proposed rectangular cross-attention (RCA) module. The features are split into vertical and horizontal windows, and the cross-attention is performed in vertical and horizontal windows separately.}\label{fig:RCA}
\end{figure}

\subsubsection{RGB-modality guided super-resolution}
\label{sec:rgsr}
The RGB-guided attention super-resolution network, i.e. RGAN, is designed to obtain high-resolution HSIs with high-resolution RGB-modality as auxiliary information. As demonstrated in Fig.~\ref{fig:srn}, RGAN is composed of multiple guided attention layers (GALs). Each GAL consists of a self-attention layer (SAL), cross-attention layer (CAL), spectral attention layer (SpecAL) and feed-forward layer (FFD). We adopt the rectangle self-attention (RA) module and spectral enhancement (SE) module proposed in~\cite{li2023spectral} as the main components of CAL and SpecAL to extract spatial and spectral correlations efficiently. FFD is composed of a normalization layer and two linear layers with a ReLU activation. In addition, inspired by the rectangle self-attention proposed in~\cite{li2023spectral}, a novel rectangular cross-attention (RCA) module is designed to aggregate the mutual contextual information between RGB and HSI efficiently.

\begin{figure}[t]
  \centering
  \includegraphics[width=0.9\linewidth]{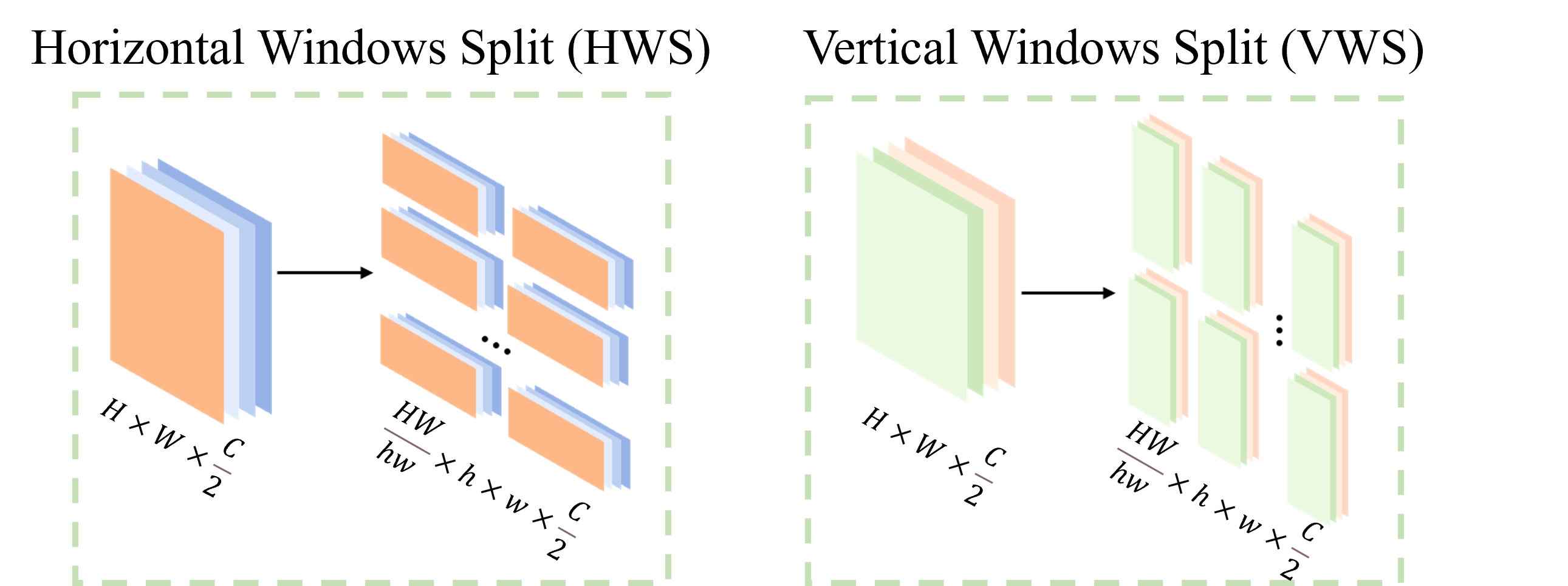}
  \caption{Illustration of rectangular windows split.}\label{fig:ws}
\end{figure}

The details of our proposed RCA module are illustrated in Fig.~\ref{fig:RCA}. The cross-attention is conducted in vertical and horizontal rectangles separately. Specifically, let $\boldsymbol{Z}_1 \in \mathbb{R}^{H\times W\times C}$ and $\boldsymbol{Z}_2 \in \mathbb{R}^{H\times W\times C}$ denote the input features, the query, key and value features of the two inputs can be calculated as
\begin{equation}\label{eq:qkv}
\begin{split}
\boldsymbol{Q}_1, \boldsymbol{K}_1, \boldsymbol{V}_1 =& \; \mathrm{Split}(\mathrm{Conv2d}(\boldsymbol{Z}_1)), \\
\boldsymbol{Q}_2, \boldsymbol{K}_2, \boldsymbol{V}_2 =& \; \mathrm{Split}(\mathrm{Conv2d}(\boldsymbol{Z}_2)),
\end{split}
\end{equation}
where $\mathrm{Conv2d}$ denotes a 2-D convolution layer, $\boldsymbol{Q}_1$,$\boldsymbol{K}_1$,$\boldsymbol{V}_1\in \mathbb{R}^{H\times W\times C}$ and $\boldsymbol{Q}_2$,$\boldsymbol{K}_2$,$\boldsymbol{V}_2\in \mathbb{R}^{H\times W\times C}$. In the following, the features are divided into two parts in spectral domain, which conduct the horizontal multi-head cross attention (H-MCA) and vertical multi-head cross attention (V-MCA) separately. Formally, the spectral split is denoted as
\begin{equation}\label{eq:ss}
\small
\begin{split}
& \boldsymbol{Q}_1^1,\boldsymbol{Q}_1^2 = \text{Split}(\boldsymbol{Q}_1), \boldsymbol{K}_1^1,\boldsymbol{K}_1^2 = \text{Split}(\boldsymbol{K}_1), \boldsymbol{V}_1^1,\boldsymbol{V}_1^2 = \text{Split}(\boldsymbol{V}_1), \\
& \boldsymbol{Q}_2^1,\boldsymbol{Q}_2^2 = \text{Split}(\boldsymbol{Q}_2), \boldsymbol{K}_2^1,\boldsymbol{K}_2^2 = \text{Split}(\boldsymbol{K}_2), \boldsymbol{V}_2^1,\boldsymbol{V}_2^2 = \text{Split}(\boldsymbol{V}_2). \\
\end{split}
\end{equation}
Then, $\boldsymbol{Q}_1^1$,$\boldsymbol{Q}_2^1$,$\boldsymbol{K}_1^1$,$\boldsymbol{K}_2^1$,$\boldsymbol{V}_1^1$,$\boldsymbol{V}_2^1$ are partitioned into horizontal windows and $\boldsymbol{Q}_1^2$,$\boldsymbol{Q}_2^2$,$\boldsymbol{K}_1^2$,$\boldsymbol{K}_2^2$,$\boldsymbol{V}_1^2$,$\boldsymbol{V}_2^2$ are partitioned into vertical windows, respectively. As shown in Fig.~\ref{fig:ws}, given the size of the rectangular window is $[h, w]$, the input features are partitioned into non-overlapping rectangular patches, and the size of each patch is $[h, w, \frac{C}{2}]$. In the case of the horizontal partition, the width $w$ is greater than the height $h$ whereas in the case of the vertical partition the height $h$ is greater than the width $w$. Next, these windows are fed into the multi-head cross-attention (MCA) module to integrate the features between the RGB modality and HSI modality. More precisely, the cross-attention is calculated as
\begin{equation}\label{eq:ca}
\begin{split}
\hat{\boldsymbol{Z}_1^1} &= \mathrm{SoftMax}(\boldsymbol{Q}_2^1\boldsymbol{K}_1^{1^\mathrm{T}}/\sqrt{d}+\boldsymbol{P})\boldsymbol{V}_1^1, \\
\hat{\boldsymbol{Z}_1^2} &= \mathrm{SoftMax}(\boldsymbol{Q}_2^2\boldsymbol{K}_1^{2^\mathrm{T}}/\sqrt{d}+\boldsymbol{P})\boldsymbol{V}_1^2, \\
\hat{\boldsymbol{Z}_2^1} &= \mathrm{SoftMax}(\boldsymbol{Q}_1^1\boldsymbol{K}_2^{1^\mathrm{T}}/\sqrt{d}+\boldsymbol{P})\boldsymbol{V}_2^1, \\
\hat{\boldsymbol{Z}_2^2} &= \mathrm{SoftMax}(\boldsymbol{Q}_1^2\boldsymbol{K}_2^{2^\mathrm{T}}/\sqrt{d}+\boldsymbol{P})\boldsymbol{V}_2^2,
\end{split}
\end{equation}
where $\boldsymbol{P}$ is the learnable position embedding and $d$ is the feature dimension.
Then the output windows are aggregated and reversed to feature maps. Finally, the outputs are obtained by concatenating the feature maps, i.e.,
\begin{equation}\label{eq:concat}
\begin{split}
\hat{\boldsymbol{Z}_1} &= \text{Concat}(\hat{\boldsymbol{Z}_1^1}, \hat{\boldsymbol{Z}_1^2}), \\
\hat{\boldsymbol{Z}_2} &= \text{Concat}(\hat{\boldsymbol{Z}_2^1}, \hat{\boldsymbol{Z}_2^2}).
\end{split}
\end{equation}
Multi-head attention mechanism is also adopted in the RCA module, which indicates that we employ several groups of parameters to conduct the cross-attention, and the results are then combined to capture a more comprehensive and informative representation of the data.

In summary, the input feature maps of HSI and RGB are divided into vertical and horizontal rectangular windows. Cross-attention is then applied between the vertical windows of the HSI and RGB feature maps, as well as between the horizontal windows. This rectangular attention mechanism enables the network to transfer the image details from the RGB modality to the HSI modality effectively and efficiently, leading to improved performance.

\begin{figure*}[t]
  \centering
  \includegraphics[width=0.7\textwidth]{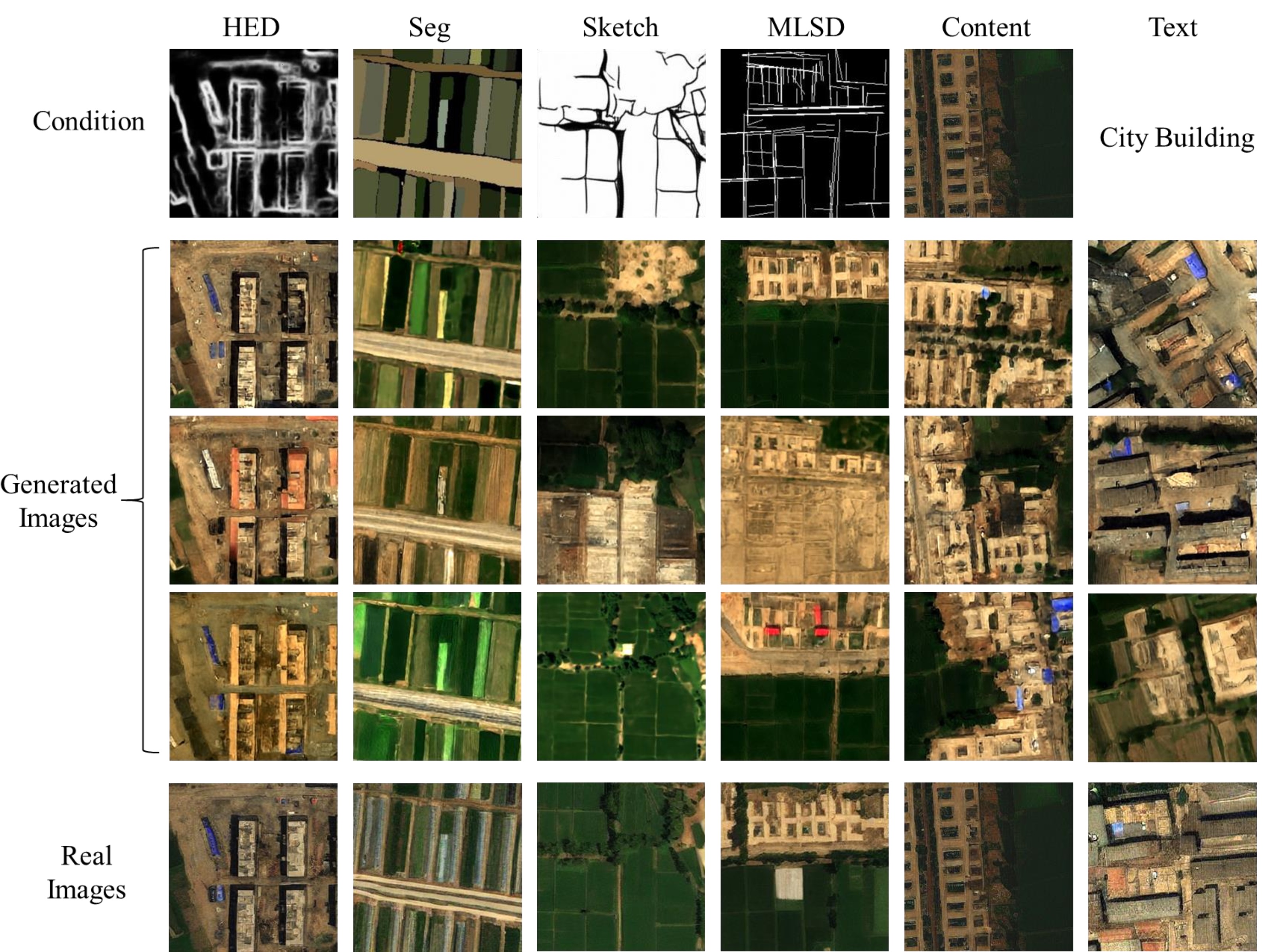}
  \caption{Visualization results of synthesized HSIs which are generated under a single condition. Our proposed HSIGene supports image generation under single conditions, such as HED, segmentation, sketch, MLSD, content and text.}\label{fig:ss}
\end{figure*}

\section{Experiments}
\label{sec:experiments}
\subsection{Experimental Settings}
\subsubsection{Synthesis experiments setting}
The datasets used for HSI synthesis are listed in Table~\ref{tab:data} and the details are illustrated in Sec.~\ref{sec:data}. Considering the spectral discrepancy between different datasets, we align their wavelength to 400-1000nm with 48 bands utilizing linear interpolation which could ensure spectral information preservation~\cite{liu2022physics}. The datasets are cropped to size $256\times 256$ with stride 128, resulting in approximately 7k training images. For data augmentation, we employ the method introduced in Sec.~\ref{sec:method} to obtain high-resolution images of the Heihe dataset with upscale factor $\times 2$ and Chikusei dataset with upscale factor $\times 4$. By cropping the super-resolved images of size $256\times 256$ with stride 128, we obtain 40k training images in total. To validate the robustness and generalization ability of our model, farmland images from the AID~\cite{xia2017aid} dataset are used for evaluation, which exhibits scenes similar to real HSIs~\cite{yu2024unmixing}. The AID dataset is also cropped with a stride of 128, resulting in 1k image patches of size $256\times 256$ in total. For each patch, we generate control conditions using the method introduced in Sec.~\ref{sec:data} which are subsequently employed to synthesize HSIs. Finally, the quality of the generated images is evaluated to assess the model's ability to generate realistic HSI data.


We use the AdamW optimizer~\cite{kingma2014adam} with a learning rate of $10^{-5}$ to train the HSI synthesis diffusion model for 100k iterations. The batch size is set as 16. Our model contains 1.5 billion parameters in total, ensuring desirable generation under scaling laws. In addition, we initialize the parameters of the UNet component from the pretrained weights~\cite{tang2024crs}, ensuring that the diffusion model adapts effectively and efficiently to HSI synthesis.

\subsubsection{Super-resolution experiments setting}
Approximately 55k high-resolution remote sensing images obtained from Google Earth Engine~\cite{gorelick2017google} are used to train the DSRNet described in Sec.\ref{sec:dbsr}. We use real HSI data (i.e., the 7k cropped image patches) and select the RGB bands as auxiliary guidance to train the RGAN model proposed in~\ref{sec:rgsr}. The training pairs are obtained by downsampling existing real HSIs with scales of 2 and 4 using the strategy~\cite{gu2019blind}.

We adopt the same training strategy of the HSI synthesis model for DSRNet. For training RGAN, the learning rate is set to $10^{-4}$ with AdamW optimizer and the learning rate finally decreases to $10^{-5}$ with a cosine annealing strategy~\cite{loshchilov2016sgdr}. The batch size is set as 2 and the total epoch number is set as 30. L1 loss is used to train the network. To evaluate the image quality of the augmented data, we randomly select 100 real images from the Heihe dataset and compare the quality of these images super-resolved by different methods. Six super-resolution methods including MCNet~\cite{li2020mixed}, SSPSR~\cite{jiang2020learning}, Bi-3DQRNN~\cite{fu2021bidirectional}, SwinIR~\cite{liang2021swinir}, ESSAformer~\cite{zhang2023essaformer}, DSTrans~\cite{yu2023dstrans} are taken for comparison. For a fair comparison, all competitive models are trained in the same setting.

\begin{figure*}[t]
  \centering
  \includegraphics[width=0.7\textwidth]{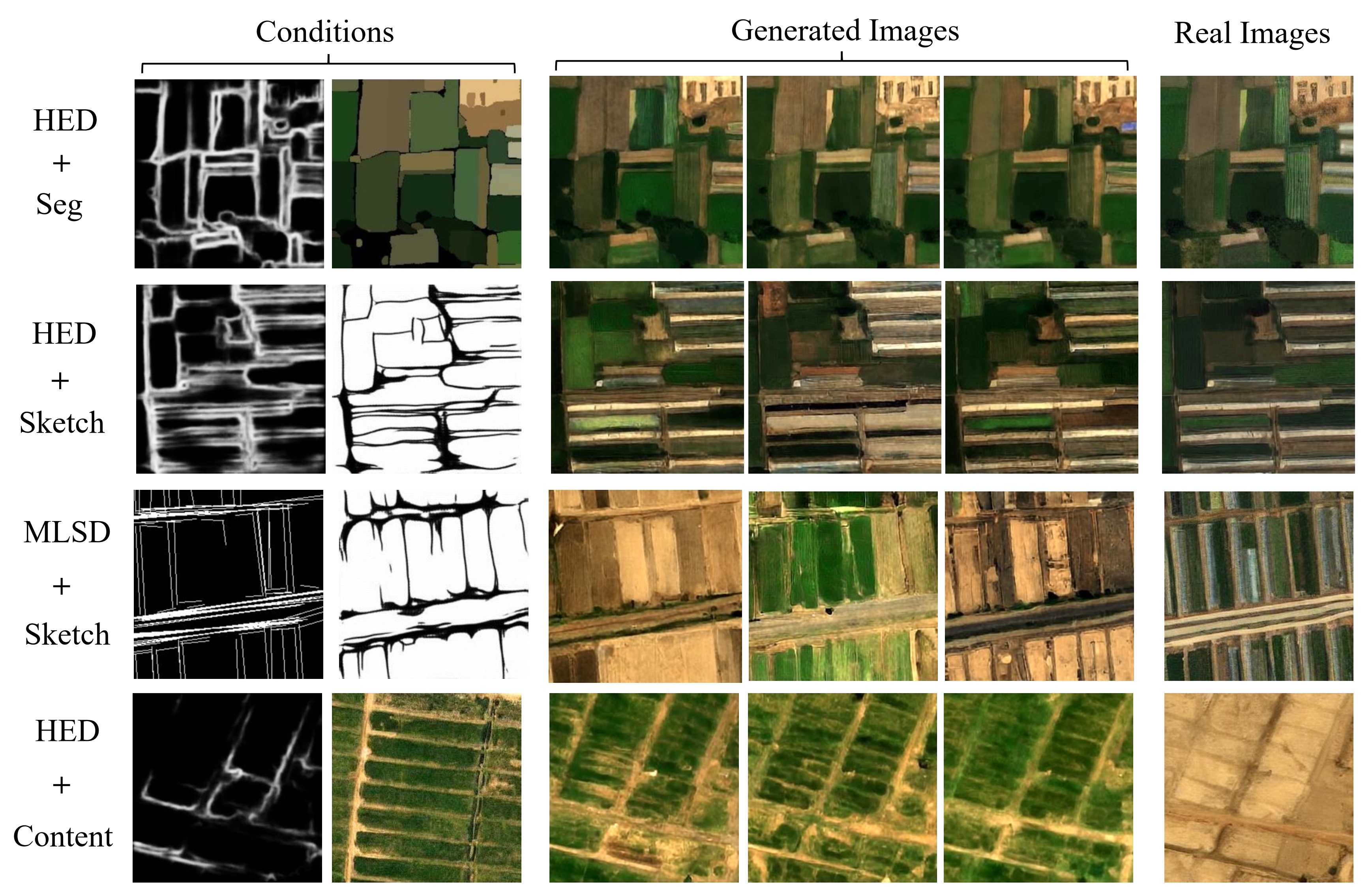}
  \caption{Visualization results of synthesized HSIs which are generated under multiple conditions. Simultaneous control of multiple conditions ensures more accurate and reliable HSI generation. }\label{fig:ms}
\end{figure*}

\begin{figure}[t]
  \centering
  \includegraphics[width=0.95\linewidth]{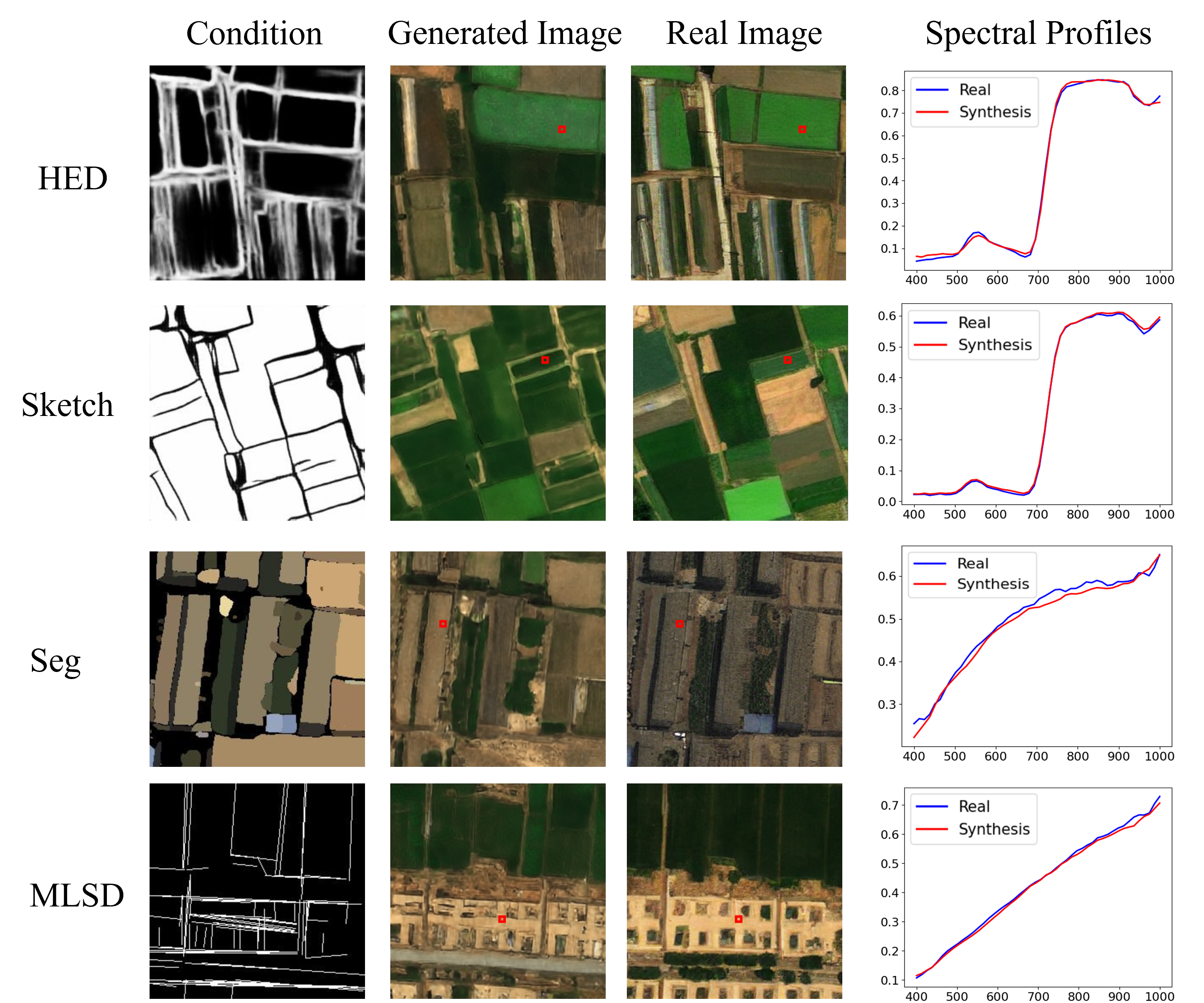}
  \caption{The comparison between generated spectral profiles and real spectral profiles. }\label{fig:spec}
\end{figure}

\begin{figure}[t]
  \centering
  \includegraphics[width=0.95\linewidth]{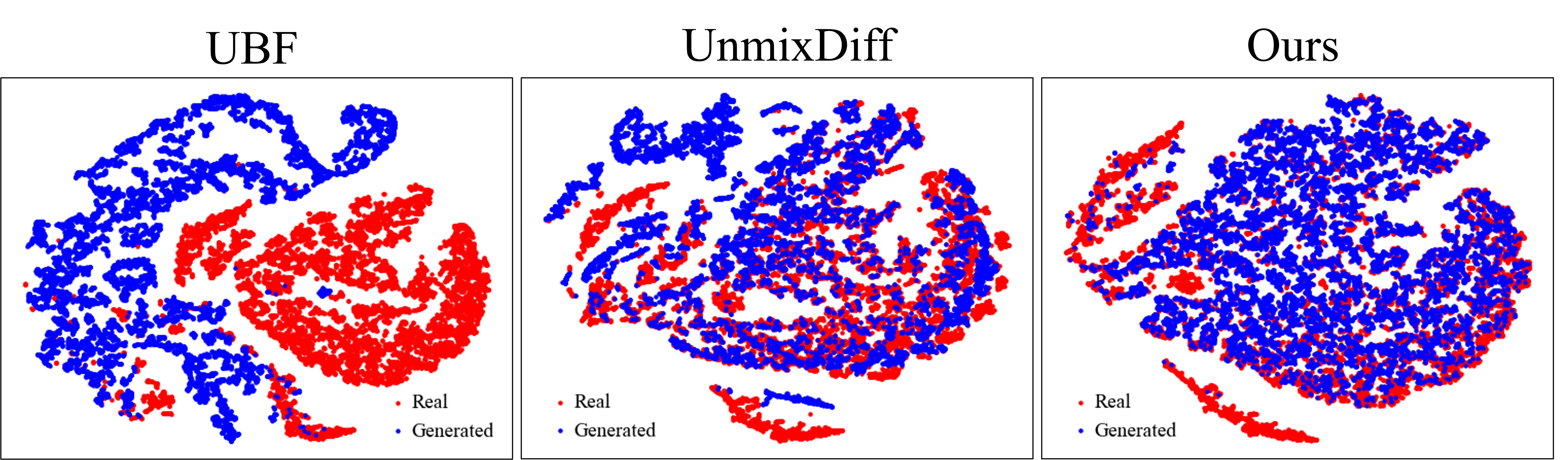}
  \caption{T-SNE visualizations of the spectra generated by UBF, UnmixDiff, and our proposed method. \textcolor{red}{Red dots} represent real spectra, while \textcolor{blue}{blue dots} represent generated spectra.}\label{fig:spec_tsne}
\end{figure}

\subsubsection{Evaluation metrics}
To assess the performance and quality of the generated images, we employ a comprehensive set of evaluation metrics, each capturing different aspects of image quality and fidelity. Inception Score (IS)~\cite{salimans2016improved} and Fr\'{e}chet Inception Distance (FID)~\cite{heusel2017gans} are used to measure the similarity between the distribution of generated HSIs and real HSIs. The CLIPScore~\cite{radford2021learning} and SSIM~\cite{wang2004image} are used to assess the similarity of content and structure respectively. Various no-reference image quality assess indexes, including NIQE~\cite{mittal2012making}, PI~\cite{blau20182018}, BRISQUE~\cite{mittal2012no}, ILNIQE~\cite{zhang2015feature}, ClipIQA~\cite{wang2023exploring} and CNNIQA~\cite{kang2014convolutional}, are utilized to provide a robust assessment of image perceptual quality. The no-reference image quality assessment is performed on the PyIQA framework~\cite{pyiqa}. In addition to the spatial evaluation, inspired by~\cite{kynkaanniemi2019improved}, we further propose two novel metrics namely Spectral Precision (sPr) and Spectral Recall (sRec) to evaluate the generated spectra. Specifically, defining real and generated spectral profiles as $\boldsymbol{\phi}_r$ and $\boldsymbol{\phi}_g$, and the corresponding spectral sets as $\boldsymbol{\Phi}_r$ and $\boldsymbol{\Phi}_g$, sPr and sRec could be calculated as follows:
\begin{equation}\label{eq:metric}
\begin{split}
\mathrm{sPr}(\boldsymbol{\Phi}_{r},\boldsymbol{\Phi}_{g})&=\frac{1}{|\boldsymbol{\Phi}_{g}|}\sum_{{\boldsymbol{\phi}_{g}\in\boldsymbol{\Phi}_{g}}}f(\boldsymbol{\phi}_{g},\boldsymbol{\Phi}_{r}) \\
\mathrm{sRec}(\boldsymbol{\Phi}_{r},\boldsymbol{\Phi}_{g})&=\frac{1}{|\boldsymbol{\Phi}_{r}|}\sum_{{\boldsymbol{\phi}_{r}\in\boldsymbol{\Phi}_{r}}}f(\boldsymbol{\phi}_{r},\boldsymbol{\Phi}_{g})
\end{split}
\end{equation}
where $f(\boldsymbol{\phi},\boldsymbol{\Phi})$ is a binary function that returns 1 if $\boldsymbol{\phi}$ is within the $k$-nearest neighbors of any $\boldsymbol{\phi}' \in \boldsymbol{\Phi}$, and otherwise returns 0. Mathematically, $f(\boldsymbol{\phi},\boldsymbol{\Phi})$ returns 1 if
\begin{equation*}\label{eq:binaryfunc}
\|\boldsymbol{\phi} - \boldsymbol{\phi}' \|_2 \leq \|\boldsymbol{\phi}' - NN_k(\boldsymbol{\phi}', \boldsymbol{\Phi})\|_2
\end{equation*}
for at least one $\boldsymbol{\phi}' \in \boldsymbol{\Phi}$, where $NN_k(\boldsymbol{\phi}', \boldsymbol{\Phi})$ returns the $k$th nearest spectral profile of $\boldsymbol{\phi}'$ from set $\boldsymbol{\Phi}$. sPr evaluates whether each generated spectral profile falls within the estimated manifold of real profiles while sRec evaluates whether each real spectral profile falls within the estimated manifold of generated profiles. In our experimental settings, we randomly sample $100, 000$ generated profiles and real profiles respectively for sPr and sRec calculation, and $k$ is set as 10. Considering the expensive computational cost of pairwise distance calculation, we divide the profiles into 10 groups. sPr and sRec are calculated for each group, and the final sPr and sRec values are obtained by averaging the results across all groups.
Among these metrics, the smaller the values of FID, NIQE, PI, ILNIQE, BRISQUE, and the larger the values of SSIM, IS, CLIPIQA, CLIPScore, sPr, sRec, the better the quality of the generated images.

\subsection{HSI synthesis results}
\subsubsection{Qualitative results}
In this section, we demonstrate the effectiveness of our model in achieving both single-condition and multi-condition control. The results of single-condition generation are illustrated in Fig.~\ref{fig:ss}. As shown in the figure, the generated images are consistent with the specified condition but also exhibit pixel-wise differences. For example, given the sketch of farmland, the generated images demonstrate highly similar structures while the image contents include buildings and wasteland. By employing conditions that indicate textural attributes, such as HED or sketch, the generated images successfully maintain the structure of the input conditions. On the other hand, when the control condition is related to the image content, the model generates images that align closely with the reference content and exhibit various structures. However, single-condition control fails to guarantee simultaneous control over both the structure and content of the generated images. In addition, the model under single-condition constraints is unable to integrate multiple forms of guidance to achieve more precise and coherent control.

The results of multi-condition generation are illustrated in Fig.~\ref{fig:ms}. Compared with single-condition generation, more conditions enable more sophisticated and accurate generation. For example, as shown in Fig.~\ref{fig:ms}, given conditions of HED and content indicating farmland, the generated images not only align with the structural prerequisites defined by the HED condition but also accurately reflect the content requirement. The multi-condition generation capability of our model offers significant advantages in producing images that are both structurally accurate and content-specific. Our model effectively handles different modalities and generates high-quality images, verifying its effectiveness and robustness to various conditions.

Additionally, the visualization comparison between the generated and real spectral profiles is provided in Fig.~\ref{fig:spec}. We select some generated HSIs whose content is similar to real HSIs, and for these images, we plot the spectral profiles at the same point for analysis. It can be seen that the generated spectral profiles are consistent with real spectral profiles, proving the effectiveness and usability of our proposed model in HSI synthesis. We also provide t-SNE~\cite{van2008visualizing} visualizations of the spectra generated by UBF~\cite{yu2024unmixing}, UnmixDiff~\cite{yu2024unmixdiff}, and our proposed method as shown in Fig.~\ref{fig:spec_tsne}. It can be seen that the distribution of spectra generated by our method covers the real distribution more effectively. UBF generates spectra with significant differences from the real spectra due to the discrepancy between RGB modality image content and HSI content. UnmixDiff, on the other hand, generates abundance maps rather than directly generating HSIs, leading to performance limitations imposed by the unmixing network. Furthermore, UnmixDiff has a smaller model size compared to our model, which restricts its generation capabilities. In contrast, the latent diffusion technique that we used enables us to train a larger model, thus leading to better performance.

\begin{table}[t]
\centering
\caption{Results comparison between our proposed HSIGene and existing HSI synthesis methods.}
\label{tab:generation}
\renewcommand\arraystretch{1}
\resizebox{\linewidth}{!}{
\begin{tabular}{lccccccc}
\toprule
          & IS\,$\uparrow$            & FID\,$\downarrow$             & NIQE\,$\downarrow$           & BRISQUE\,$\downarrow$         & ClipIQA\,$\uparrow$        & sPr\,$\uparrow$            & sRec\,$\uparrow$        \\ \midrule
UBF~\cite{yu2024unmixing}       & 1.091          & 123.377         & 8.269          & 34.906          & 0.371          & 0.399          & 0.113         \\
UnmixDiff~\cite{yu2024unmixdiff} & 1.180          & 111.281         & \textbf{6.339} & 33.552          & 0.417          & 0.758          & 0.573         \\
Ours      & \textbf{1.200} & \textbf{76.073} & 6.447          & \textbf{31.622} & \textbf{0.456} & \textbf{0.988} & \textbf{0.846} \\ \bottomrule
\end{tabular}
}
\end{table}

\begin{table}[t]
\centering
\caption{Comparison of image generation results under different combinations of control conditions. The best results are in \textbf{bold}, and the second-best results are \underline{underlined}.}
\label{tab:multicond}
\renewcommand\arraystretch{1}
\resizebox{\linewidth}{!}{
\begin{tabular}{lccccccc}
\toprule
Conditions          & SSIM\,$\uparrow$           & CLIPScore\,$\uparrow$       & NIQE\,$\downarrow$           & BRISQUE\,$\downarrow$         & ClipIQA\,$\uparrow$           & sPr\,$\uparrow$        & sRec\,$\uparrow$      \\ \midrule
MLSD                & 0.267          & 79.273          & 6.312          & 34.670          & 0.474              & 0.872     & \textbf{0.975}   \\
MLSD+HED            & 0.359          & \underline{81.522}    & \underline{6.015}    & 27.352          & \textbf{0.573}  & \underline{0.992} & 0.914   \\
MLSD+HED+Sketch     & \underline{0.362}     & 81.398          & \textbf{6.013} & \underline{27.071}    & \underline{0.569}     & 0.990 & \underline{0.931}     \\
MLSD+HED+Sketch+Seg & \textbf{0.370} & \textbf{81.832} & 6.109          & \textbf{26.430} & \textbf{0.573}  & \textbf{0.993} & 0.852  \\ \bottomrule
\end{tabular}
}
\end{table}

\begin{table*}[t]
\centering
\caption{Results on the AID dataset when models are trained with/without augmented data.}
\label{tab:aug2}
\renewcommand\arraystretch{1.0}
\resizebox{0.7\textwidth}{!}{
\footnotesize
\begin{tabular}{llccccccc}
\toprule
                        &         & SSIM\,$\uparrow$           & ClipScore\,$\uparrow$ & NIQE\,$\downarrow$           & BRISQUE\,$\downarrow$         & ClipIQA\,$\uparrow$        & sPrn\,$\uparrow$     & sRec\,$\uparrow$        \\ \hline
\multirow{2}{*}{HED}    & w/o aug & 0.349          & -         & 6.197          & 36.898          & 0.415          & 0.946          & 0.938          \\
                        & Ours    & \textbf{0.359} & -         & \textbf{6.037} & \textbf{27.125} & \textbf{0.571} & \textbf{0.966} & \textbf{0.946} \\ \hline
\multirow{2}{*}{MLSD}   & w/o aug & 0.225          & -         & 6.722          & 40.546          & 0.419          & \textbf{0.899} & 0.925          \\
                        & Ours    & \textbf{0.267} & -         & \textbf{6.312} & \textbf{34.670} & \textbf{0.474} & 0.872          & \textbf{0.975} \\ \hline
\multirow{2}{*}{Sketch} & w/o aug & 0.216          & -         & 7.400          & 41.585          & 0.417          & \textbf{0.878} & 0.918          \\
                        & Ours    & \textbf{0.265} & -         & \textbf{6.379} & \textbf{32.823} & \textbf{0.484} & 0.871          & \textbf{0.970} \\ \hline
\multirow{2}{*}{Seg}    & w/o aug & 0.283          & -         & 6.449          & 39.170          & 0.424          & \textbf{0.932} & 0.877          \\
                        & Ours    & \textbf{0.310} & -         & \textbf{6.502} & \textbf{31.039} & \textbf{0.511} & 0.911          & \textbf{0.965} \\ \hline
\multirow{2}{*}{Content} & w/o aug & -              & 77.239          & \textbf{7.262} & 41.066          & 0.396          & 0.945          & 0.918          \\
                         & Ours    & -              & \textbf{78.736} & 7.268          & \textbf{32.261} & \textbf{0.459} & \textbf{0.962} & \textbf{0.934} \\ \hline
\multirow{2}{*}{Text}    & w/o aug & -              & 22.465          & \textbf{6.741} & 35.476          & \textbf{0.410} & \textbf{0.995} & 0.770          \\
                         & Ours    & -              & \textbf{22.752} & 6.962          & \textbf{34.256} & 0.408          & 0.988          & \textbf{0.836} \\ \bottomrule
\end{tabular}
}
\end{table*}

\subsubsection{Quantitative results}
We comprehensively compare our model with existing HSI synthesis models, i.e., UBF~\cite{yu2024unmixing} and UnmixDiff~\cite{yu2024unmixdiff}. For fairness, the evaluation is performed on 1024 images generated by each model in an unconditional setting. The results of this comparison are presented in Table~\ref{tab:generation}, where our approach demonstrates superior performance on most metrics. UBF trains a diffusion model on augmented training datasets which is obtained by performing spectral super-resolution on RGB images. Owing to the inherent differences between RGB images and hyperspectral data, while UBF is capable of generating HSIs, it is challenging to ensure the spectral authenticity of the generated images, resulting in both low sPr and sRec. UnmixDiff, on the other hand, is trained directly on real HSI data. However, its effectiveness is limited by the relatively small size of available hyperspectral datasets and model parameters, restricting its ability to generalize and produce high-quality images across diverse scenarios. In addition, the reliability of the generated spectra is constrained by the performance of the unmixing network in UnmixDiff, which generates abundance maps instead of HSIs directly. Our model addresses these limitations by increasing both model size and training data based on latent diffusion and spatial super-resolution augmentation. Particularly, the spatial super-resolution approach significantly increases the number of data samples while maintaining the spectral reliability of the images. This method not only enhances the diversity of the training data but also ensures that the generated images preserve the spectral characteristics of real hyperspectral data, leading to the best performance among the compared models. In addition, the latent representation generation technique adopted in our model enables a larger model size, leading to improved generation performance under scaling laws.

Besides, compared with UBF~\cite{yu2024unmixing} and UnmixDiff~\cite{yu2024unmixdiff}, our model supports HSI generation under multiple conditions, enabling more accurate and effective generation. Image generation results with the conditions generated by the AID dataset are shown in Table~\ref{tab:multicond}. As can be seen, our model obtains superior performance when more conditions are provided. For example, when four conditions (i.e., MLSD, HED, sketch and segmentation) are simultaneously applied, our model achieves the best results across metrics including SSIM, CLIPScore, BRISQUE and sPr. The results indicate that additional control inputs allow the model to better guide the generation process, resulting in images that are not only structurally accurate but also of higher fidelity and perceptual quality.

\subsection{Augmentation analysis}
In this section, we provide an analysis of our proposed HSI augmentation method. We verify the effectiveness of our proposed HSI augmentation method in two aspects, including the improvement in HSI generation and the image quality of augmented data.

\subsubsection{Improvement in HSI generation}
The effectiveness of our data augmentation method is demonstrated through a comparison of models trained with and without data augmentation, as shown in Table~\ref{tab:aug2}. The results indicate that the model trained with augmented data achieves superior performance in most cases, verifying the effectiveness of our proposed data augmentation strategy. As the number of training samples increases, the model is trained on a more diverse set of conditions, leading to improved generalization performance and the ability to generate images that align with various control conditions. Through the application of diffusion models (i.e., DSRNet), we achieve data augmentation while preserving the high spatial quality of the data. The model trained with augmented data shows slightly lower performance in sPr but higher sRec in some cases. This outcome is intuitive because the resolution of the augmented data is higher than that of the real dataset. As a result, the model trained on augmented data is more likely to generate images with higher resolution than the original HSIs, which can lead to more various spectral profiles comparing to real hyperspectral images. Since spectral evaluation metrics (i.e., sPr and sRec) are calculated between generated and real spectra, and the ground truth spectra of high-resolution images are unknown, the sPr metric of spectral evaluation may be slightly lower than models trained without super-resolved data. In contrast, due to the increased spectral diversity, the sRec metric tends to be higher. Despite this, the model achieves a good balance between sPr and sRec, ensuring the authenticity of the generated spectra. In addition, the model trained with augmented data exhibits better performance in both structural similarity and overall perceptual image quality, demonstrating that data augmentation not only enhances the model's ability to generalize across different conditions but also enhances its robustness and effectiveness in practical applications.

\begin{figure}[t]
  \centering
  \includegraphics[width=0.9\linewidth]{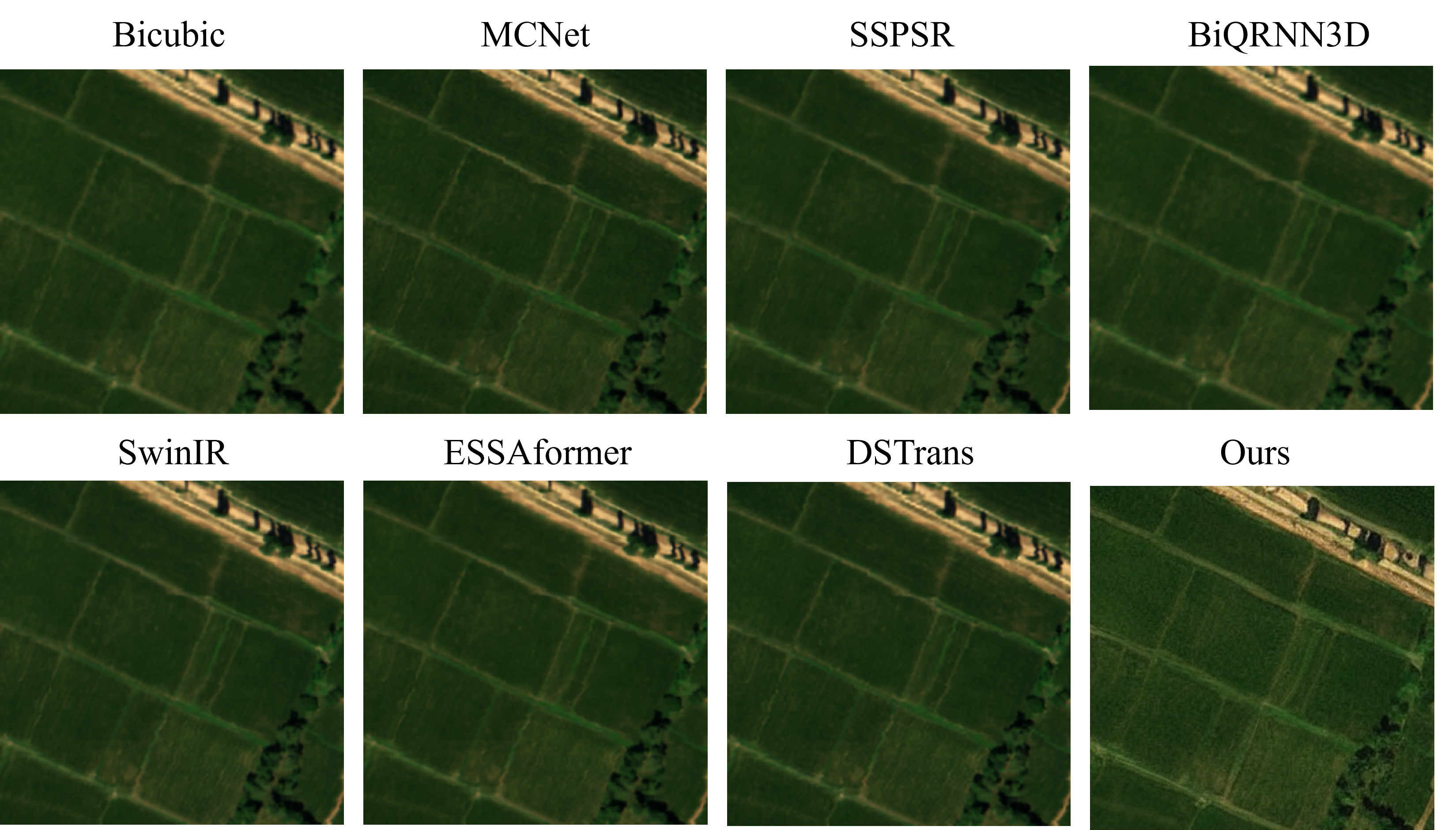}
  \caption{Visualization comparison of the augmented data super-resolved by different approaches. }\label{fig:sr}
\end{figure}

\begin{table*}[t]
\centering
\caption{Comparison of the quality of augmented data produced by different methods.}
\label{tab:sr}
\renewcommand\arraystretch{1.1}
\footnotesize
\begin{tabular}{lcccccccc}
\toprule
           & NIQE\,$\downarrow$           & PI\,$\downarrow$             & BRISQUE\,$\downarrow$         & ILNIQE\,$\downarrow$          & ClipIQA\,$\uparrow$        & CNNIQA\,$\uparrow$         & sPr\,$\uparrow$     & sRec\,$\uparrow$        \\ \midrule
Bicubic    & 7.513          & 7.033          & 45.396          & 89.439          & 0.386          & 0.200          & 0.989          & 0.835          \\
MCNet~\cite{li2020mixed}      & 6.654          & 6.314          & 36.817          & 73.177          & 0.458          & 0.276          & 0.987          & 0.840          \\
SSPSR~\cite{jiang2020learning}      & 7.303          & 6.826          & 41.455          & 76.972          & 0.405          & 0.240          & 0.990          & 0.833          \\
BIQRNN3D~\cite{fu2021bidirectional}   & 7.362          & 6.859          & 42.043          & 78.250          & 0.390          & 0.242          & 0.989          & 0.836          \\
SwinIR~\cite{liang2021swinir}     & 7.354          & 6.810          & 40.751          & 75.551          & 0.407          & 0.242          & \textbf{0.992} & 0.818          \\
ESSAformer~\cite{zhang2023essaformer} & 7.093          & 6.654          & 40.335          & 74.256          & 0.408          & 0.263          & 0.991          & 0.831          \\
DSTrans~\cite{yu2023dstrans}    & 6.885          & 6.511          & 39.719          & 75.456          & 0.434          & 0.262          & \textbf{0.992} & 0.821          \\
Ours       & \textbf{6.121} & \textbf{5.695} & \textbf{15.572} & \textbf{60.075} & \textbf{0.490} & \textbf{0.387} & 0.963          & \textbf{0.868} \\ \bottomrule
\end{tabular}
\end{table*}


\subsubsection{Image quality of augmented data}
To measure the image quality of the augmented data and validate the effectiveness of our super-resolution method, we provide quantitative results of no-reference image quality assessment metrics in Table~\ref{tab:sr} and visual results in Fig.~\ref{fig:sr}. It can be seen that our method significantly outperforms other approaches in image perceptual quality. Our method is able to achieve superior results for several reasons. First, the diffusion model (i.e., DSRNet) adopted in our network could effectively recover image details, which is crucial for high-quality super-resolution. Second, we leverage a large dataset of high-resolution RGB images for training, significantly enhancing the network’s performance by providing a rich source of detail and structure. Moreover, the RGAN network proposed in our work effectively transfers the fine details from the RGB modality into the hyperspectral modality, ensuring that the super-resolved hyperspectral images retain high fidelity and sharpness. Owing to the generative ability of DSRNet, the super-resolved images exhibit richer high-frequency details, leading to more diverse spectral profiles. Therefore, as discussed in the previous section, our model performs better in terms of sRec but slightly worse in terms of sPr. Despite this, our model achieves desirable results on both sPr and sRec metrics, confirming the spectral reliability of the augmented data.

\subsection{Ablation results}
In this section, we discuss the effectiveness of the components of our proposed two-stage super-resolution method. We compare our approach to two alternative situations: directly applying the diffusion model DSRNet proposed in Sec.\ref{sec:dbsr} or the RGAN network proposed in Sec.\ref{sec:rgsr} for HSI super-resolution and training these networks on real HSI data. For the DSRNet we directly employ the low-resolution HSIs as conditions and for the RGAN the input of RGB-modality is set as zeros. The results are illustrated in Table~\ref{tab:component}. It can be seen that, DSRNet and RGAN, when used independently, struggle to recover high-frequency details in the images. In contrast, our two-stage method leverages both high-resolution data augmentation and the powerful generative capabilities of the diffusion model. The first stage uses high-resolution RGB images to enhance the training dataset, ensuring that the diffusion model learns to reproduce intricate details of high-resolution RGB bands. In the second stage, image details are transferred to HSIs using the RGAN, resulting in superior super-resolution performance. Overall, by combining the strengths of high-resolution data augmentation with the diffusion model’s generative power, our approach effectively recovers high-frequency details and produces the highest quality super-resolved HSIs.

\begin{table}[t]
\centering
\caption{Ablation study of the components of the super-resolution framework.}
\label{tab:component}
\renewcommand\arraystretch{1.1}
\resizebox{1\linewidth}{!}{
\begin{tabular}{lcccccccc}
\toprule
              & NIQE\,$\downarrow$           & PI\,$\downarrow$             & BRISQUE\,$\downarrow$         & CNNIQA\,$\uparrow$         & ClipIQA\,$\uparrow$        & sPr\,$\uparrow$     & sRec\,$\uparrow$        \\ \midrule
Ours (DSRNet) & 6.668          & 5.971          & 37.120          & 0.436          & 0.363          & 0.967          & 0.831          \\
Ours (RGAN)   & 6.234          & 5.890          & 35.338          & 0.476          & 0.334          & \textbf{0.987} & 0.846          \\
Ours          & \textbf{6.121} & \textbf{5.695} & \textbf{15.572} & \textbf{0.490} & \textbf{0.387} & 0.963          & \textbf{0.868} \\ \bottomrule
\end{tabular}
}
\end{table}

\begin{table}[h]
\centering
\caption{The HSI denoising performance of different methods with/without synthesized HSIs.}
\label{tab:down1}
\renewcommand\arraystretch{1.1}
\small
\resizebox{1\linewidth}{!}{
\begin{tabular}{l|l|cccccc}
\hline
                       &                  & \multicolumn{3}{c}{Xiongan}                       & \multicolumn{3}{c}{WHU-Hi-HanChuan}                        \\ \hline
                       & Training Data    & PSNR\,$\uparrow$            & SSIM\,$\uparrow$           & SAM\,$\uparrow$            & PSNR\,$\uparrow$            & SSIM\,$\uparrow$           & SAM\,$\uparrow$            \\ \hline
\multirow{3}{*}{T3SC~\cite{bodrito2021trainable}}  & Real             & 37.932          & \textbf{0.923} & \textbf{0.028} & 34.625          & \textbf{0.831} & 0.365          \\
                       & Synthetic        & 34.051          & 0.854          & 0.046          & 33.407          & 0.784          & 0.406          \\
                       & Real + Synthetic & \textbf{38.227} & \textbf{0.923} & \textbf{0.028} & \textbf{34.943} & 0.830          & \textbf{0.344} \\ \hline
\multirow{3}{*}{TRQ3D~\cite{pang2022trq3dnet}} & Real             & 37.695          & \textbf{0.924} & \textbf{0.027} & 24.355          & 0.491          & 0.511          \\
                       & Synthetic        & 30.570          & 0.854          & 0.122          & 31.064          & 0.776          & 0.379          \\
                       & Real + Synthetic & \textbf{38.025} & 0.923          & 0.028          & \textbf{32.362} & \textbf{0.815} & \textbf{0.370} \\ \hline
\multirow{3}{*}{SERT~\cite{li2023spectral}}  & Real             & \textbf{38.941} & \textbf{0.928} & \textbf{0.027} & 29.122          & 0.703          & 0.469          \\
                       & Synthetic        & 35.296          & 0.895          & 0.041          & 33.948          & 0.868          & 0.268          \\
                       & Real + Synthetic & 38.725          & 0.927          & \textbf{0.027} & \textbf{34.762} & \textbf{0.879} & \textbf{0.248} \\ \hline
\end{tabular}
}
\end{table}

\begin{table}[h]
\centering
\caption{The HSI super-resolution performance of different methods with/without synthesized HSIs.}
\label{tab:down2}
\renewcommand\arraystretch{1.1}
\small
\resizebox{1\linewidth}{!}{
\begin{tabular}{l|l|cccccc}
\hline
                            &                  & \multicolumn{3}{c}{Xiongan}                       & \multicolumn{3}{c}{WHU-Hi-HanChuan}                        \\ \hline
                            & Training Data    & PSNR\,$\uparrow$            & SSIM\,$\uparrow$           & SAM\,$\uparrow$            & PSNR\,$\uparrow$            & SSIM\,$\uparrow$           & SAM\,$\uparrow$            \\ \hline
\multirow{3}{*}{MCNet~\cite{li2020mixed}}      & Real             & 35.096          & 0.865          & \textbf{0.028} & 35.490          & 0.885          & 0.172          \\
                            & Synthetic        & 34.662          & 0.856          & 0.030          & 36.758          & 0.927          & 0.136          \\
                            & Real + Synthetic & \textbf{35.302} & \textbf{0.866} & \textbf{0.028} & \textbf{36.954} & \textbf{0.930} & \textbf{0.132} \\ \hline
\multirow{3}{*}{SSPSR~\cite{jiang2020learning}}      & Real             & \textbf{37.485} & 0.870          & \textbf{0.027} & 28.982          & 0.750          & 0.372          \\
                            & Synthetic        & 33.832          & 0.843          & 0.034          & 34.801          & \textbf{0.898} & 0.183          \\
                            & Real + Synthetic & 37.474          & \textbf{0.871} & \textbf{0.027} & \textbf{34.850} & \textbf{0.898} & \textbf{0.171} \\ \hline
\multirow{3}{*}{ESSAformer~\cite{zhang2023essaformer}} & Real             & \textbf{37.452} & \textbf{0.877} & \textbf{0.027} & 29.662          & 0.744          & 0.397          \\
                            & Synthetic        & 31.973          & 0.838          & 0.074          & 33.328          & 0.867          & \textbf{0.240}          \\
                            & Real + Synthetic & 37.324          & 0.872          & \textbf{0.027} & \textbf{33.845} & \textbf{0.869} & 0.250          \\ \hline
\end{tabular}
}
\end{table}

\subsection{Application in downstream tasks}
We further validate the effectiveness of our generative model on two downstream tasks: HSI denoising and HSI super-resolution. We augment the training datasets for both tasks using our proposed generative model to improve the performance of existing image restoration models. Test images are degraded with Gaussian noise ($\sigma=0.2$) for denoising and downsampled $4\times$ for super-resolution respectively. The Xiongan dataset is partitioned into two parts: one part of size $512\times 512$ for testing, and the remaining part for training. In addition, the image restoration models are also tested on the WHU-Hi-HanChuan dataset~\cite{zhong2020whu} which is not used for training our generative model. For training, the Xiongan dataset is cropped into patches whose size is $128\times 128$ pixels with a stride of 32, resulting in approximately 5k training patches. In the augmented data scenario, we used our generative model to synthesize 100 additional HSIs with a resolution of $256\times 256$. These generated images are then also cropped into patches of the size $128\times128$, producing another 5k training patches. This augmentation effectively enhances the training data, providing a total of 10k patches for training. All training data are degraded in the same setting as test images to generate pair-wise training images. We evaluate the performance using three metrics including Peak Signal-to-Noise Ratio (PSNR), Structural Similarity Index Measure (SSIM) and Spectral Angle Mapper (SAM). We employ T3SC~\cite{bodrito2021trainable}, TRQ3D~\cite{pang2022trq3dnet}, SERT~\cite{li2023spectral} for denoising evaluation and MCNet~\cite{li2020mixed}, SSPSR~\cite{jiang2020learning}, ESSAformer~\cite{zhang2023essaformer} for super-resolution evaluation. All models are trained for 30 epochs and the batch size is 4. The learning rate is $10^{-4}$ with AdamW optimizer and decreases to $10^{-5}$ with a strategy of cosine annealing.

The results are presented in Table~\ref{tab:down1} and Table~\ref{tab:down2}. As can be seen, when the test set and training set have the same distribution (i.e., the Xiongan dataset), models trained with both augmented and real data exhibit comparable performance to the model trained with only real data. However, when tested on the WHU-Hi-HanChuan dataset, the performance of models trained solely on real data, such as TRQ3D and SERT, deteriorates significantly, while the models trained with both augmented and real data still achieve the best results. Additionally, the model trained solely on augmented data also achieves relatively desirable performance, proving the high quality of our generated data. By augmenting the training data, our approach contributes to more robust and accurate models, demonstrating the broader impact and effectiveness of our work in advancing more HSI processing tasks.

\section{Conclusion}
\label{sec:conclusion}
In this paper, we present a foundation model namely HSIGene for HSI synthesis, which utilizes a latent diffusion model to synthesize HSIs with support for multiple control conditions. To enhance the spatial diversity of the training dataset, we propose a spatial super-resolution based data augmentation method and design a two-stage super-resolution approach to improve the perceptual quality of the augmented images. Extensive experiments demonstrate that our model outperforms existing methods in HSI synthesis and verifies the reliability of the augmented data. The results of two downstream tasks including HSI denoising and HSI super-resolution demonstrate that our model could provide a substantial amount of high-quality data for model training, boosting the performance of downstream tasks.


%

%

\section*{Acknowledgment}
The Heihe datasets is provided by National Tibetan Plateau / Third Pole Environment Data Center (http://data.tpdc.ac.cn). In addition, we would like to thank the Hyperspectral Image Analysis group and the NSF Funded Center for Airborne Laser Mapping (NCALM) at the University of Houston for providing the DFC data sets used in this study, and the IEEE GRSS Data Fusion Technical Committee for organizing the 2013 and 2018 Data Fusion Contest.

\ifCLASSOPTIONcaptionsoff
  \newpage
\fi



%
\ifCLASSOPTIONcaptionsoff
\newpage
\fi
\bibliographystyle{ieeetr}
\bibliography{sample-base}




\end{document}